\documentclass[journal]{IEEEtran}

\usepackage{color, array, amsthm}
\usepackage{graphicx}
\usepackage{amssymb}
\usepackage{xcolor}

\usepackage{multirow}
\usepackage[caption=false,font=footnotesize,labelfont=sf,textfont=sf]{subfig}

\usepackage{algorithm}
\usepackage{algpseudocode}
\usepackage{amsmath}
\usepackage{hyperref}
\hypersetup{
    pdfborder={0 0 0}  
}

\usepackage{adjustbox} 
\usepackage{tikz}
\usepackage{mwe} 
\usetikzlibrary{shapes, arrows, positioning, shapes.geometric, fit, arrows.meta, calc, backgrounds}
\tikzset{
  img/.style   ={draw,rounded corners=2pt,minimum width=3.6cm,minimum height=2.2cm,inner sep=0pt},
  module/.style={draw,rounded corners=3pt,minimum height=1cm,minimum width=1.5cm,align=center,font=\small},
  fe/.style={
    draw=black,
    thick,
    fill=green!20,
    align=center,
    inner sep=2pt,
    minimum width=1.2cm, 
    minimum height=1.8cm,
    shape=trapezium,
    trapezium left angle=70,
    trapezium right angle=70,
    trapezium stretches=true
  },
  stff/.style  ={module,fill=blue!15},
  sppf/.style  ={module,fill=violet!15},
  c2psa/.style ={module,fill=violet!25},
  neck/.style  ={module,fill=yellow!25},
  head/.style  ={module,fill=yellow!35},
  flow/.style  ={-Latex,thick},
  bigbox/.style={draw,rounded corners=12pt,fill=orange!8,thick,inner sep=12pt}
}

\begin{document}

\title{GL-DT: Multi-UAV Detection and Tracking with Global-Local Integration}

\author{Juanqin Liu, Leonardo Plotegher, Eloy Roura and Shaoming He\textsuperscript{*}
\thanks{This work was supported by the Technology Innovation Institute under Contract No. TII/ARRC/2154/2023.}
\thanks{Juanqin Liu and Shaoming~He are with the School of Aerospace Engineering, Beijing Institute of Technology, Beijing 100081, China.}
\thanks{Leonardo Plotegher and Eloy Roura are with the Autonomous Robotics Research Centre, Technology Innovation Institute, P.O.Box: 9639, Masdar City, Abu Dhabi, United Arab Emirates.}
\thanks{\textsuperscript{*}Corresponding Author. Email: \texttt{shaoming.he@bit.edu.cn}.}
}

\maketitle

\begin{abstract}
The extensive application of unmanned aerial vehicles (UAVs) in military reconnaissance, environmental monitoring, and related domains has created an urgent need for accurate and efficient multi-object tracking (MOT) technologies, which are also essential for UAV situational awareness. However, complex backgrounds, small-scale targets, and frequent occlusions and interactions continue to challenge existing methods in terms of detection accuracy and trajectory continuity. To address these issues, this paper proposes the Global-Local Detection and Tracking (GL-DT) framework. It employs a Spatio-Temporal Feature Fusion (STFF) module to jointly model motion and appearance features, combined with a global-local collaborative detection strategy, effectively enhancing small-target detection. Building upon this, the JPTrack tracking algorithm is introduced to mitigate common issues such as ID switches and trajectory fragmentation. Experimental results demonstrate that the proposed approach significantly improves the continuity and stability of MOT while maintaining real-time performance, providing strong support for the advancement of UAV detection and tracking technologies.
\end{abstract}


\section{Introduction}
The rapid advancement of UAV technology has established drones as indispensable tools in a wide range of applications, including military reconnaissance, environmental monitoring, and disaster response \cite{zheng2021air, isaac2021unmanned}. However, the surging number of UAVs in airspace has raised serious safety concerns, such as unauthorized flights and mid-air collisions, posing significant risks to public security \cite{Wang2025FocusTrack, guo2024global, xie2020adaptive,wang2024survey}. Consequently, achieving efficient and accurate aerial drone detection and tracking their trajectories has become a critical research focus for mitigating potential security threats.

\begin{figure}[htbp]
	\centering
	\includegraphics[width=\linewidth]{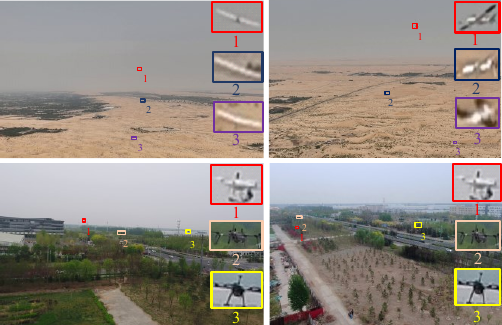}	
	\caption{Some typical examples of UAVs: the first row shows small, low-visibility UAV targets, while the second row illustrates UAVs in complex backgrounds, where targets are difficult to distinguish from the environment.}
	\label{Fig1}	
\end{figure}

Against this backdrop, UAV detection and tracking research centers on two core application scenarios. The first involves ground-based surveillance systems \cite{sun2023enhancing, lo2021dynamic}. These systems prioritize long-range, high-precision detection of small aerial targets under complex backgrounds. The second pertains to the on-board sense-and-avoid system \cite{fiorio2025development, wang2022low}. These systems require efficient, low-latency target recognition and dynamic evasion under constrained computational resources. Achieving accurate detection and continuous tracking of UAVs in such challenging environments imposes stringent requirements on algorithmic precision, robustness, and real-time performance. However, existing methods still face numerous challenges. On the one hand, UAVs are typically small, structurally simple, and exhibit weak textures, making them difficult to detect in complex aerial scenes. This challenge is exacerbated by factors such as high-speed motion, occlusion, cluttered backgrounds, lighting variations, and adverse weather conditions, under which traditional detection methods often struggle to balance accuracy and real-time performance \cite{li2024learning, liu2025small, hu2024anti}. On the other hand, in multi-UAV scenarios, targets often share highly similar visual features, while their trajectories and behavioral patterns tend to be irregular \cite{dong2025securing, zhao20233rd}. This greatly complicates identity preservation and temporal association in MOT, posing significant challenges for algorithm design and optimization. Fig. \ref{Fig1} presents some typical examples of UAVs.

In recent years, research on UAV detection and tracking has advanced considerably, with numerous vision-based solutions proposed. In detection, efforts have focused on improving the accuracy and speed of small-object recognition through novel network architectures and algorithms. For example, deep learning-based frameworks, such as You Only Look Once (YOLO), have also seen broad application in UAV detection tasks \cite{Wan2025QHNet,zhang2024domain,zhou2023admnet,aydin2023drone,guo2024motion}. Techniques such as Feature Pyramid Networks (FPN), multi-scale feature fusion, and attention mechanisms have also been widely adopted to enhance detection performance \cite{han2023light,zhao2024lightweight,reis2023real}. More recently, transformer-based models have been introduced into aerial scenarios, demonstrating strong representation and modeling capabilities \cite{meinhardt2022trackformer, zeng2022motr}. For tracking, researchers have primarily employed two paradigms: the two-stage `tracking-by-detection' approach \cite{zhang2022bytetrack, wojke2017simple, cao2023observation, aharon2022bot} and the end-to-end `joint detection and tracking' framework \cite{zhang2021fairmot, zhou2020tracking}, often in conjunction with Kalman filters and data association algorithms, to achieve accurate and efficient multi-UAV tracking in dynamic environments. Nevertheless, existing methods continue to face significant challenges, particularly in terms of adaptability, generalization across complex conditions, and real-time performance.

This paper focuses on UAV target detection and tracking, proposing a method that integrates motion and appearance features. By leveraging joint information extraction across consecutive frames and adopting a global–local collaborative detection strategy, the detection accuracy of UAV targets is significantly improved. Building upon this, we further introduce the JPTrack tracking algorithm, which effectively mitigates common issues in traditional methods such as ID switches and trajectory fragmentation. The proposed approach not only maintains real-time performance but also substantially enhances the continuity and stability of multi-object tracking. The main contributions of this work are:
\begin{itemize}
\item Design of a global–local collaborative detection framework: We propose the STFF module, which effectively fuses motion and appearance features to enhance global detection performance. By adaptively switching between global and local detection, the framework significantly improves both the accuracy and robustness of small UAV detection.

\item Development of the JPTrack algorithm: JPTrack leverages the JCMA strategy for high-precision target association and integrates the PMR module to recover trajectories under short-term occlusions or temporary target loss. This design substantially reduces ID switches and trajectory fragmentation, while ensuring stability and continuity in complex UAV scenarios.  

\item Construction of the FT Dataset: Images of 1–3 UAVs were captured from long-range perspectives in fixed-wing UAV scenarios, emphasizing low-resolution small targets (average pixel occupancy below 0.1\%), with a total of 25,855 frames. The dataset provides a benchmark and essential data support for UAV detection and multi-object tracking tasks.
\end{itemize}

\section{Related Work}
\subsection{UAV Detection Methods}
Compared to conventional object detection, UAV targets present unique challenges due to their small size, simple structure, and weak texture, often appearing as small objects in real-world imagery. Detection is further complicated by environmental clutter, background interference, and varying lighting conditions. Existing approaches primarily focus on two directions: feature enhancement and temporal information exploitation.
Feature-enhanced single-frame detection methods aim to improve local feature representation by refining deep neural networks. For instance, Han et al. \cite{han2023light} integrated Transformer blocks and a parallel mixed efficient attention module into YOLOv5 to improve regional feature extraction, while employing lightweight convolution modules to balance accuracy and efficiency. Zhou et al. \cite{zhou2023admnet} introduced a spatial attention module into the backbone and a spatial pyramid pooling (SPP) module into the prediction head, achieving multi-scale feature fusion and enhanced global context perception. While these approaches improve local feature perception, they often lack global context modeling, limiting their effectiveness in detecting ultra-small UAVs with sparse visual cues. Temporal and motion-aware methods enhance detection stability by leveraging motion cues across consecutive frames. Typical strategies include optical-flow-guided feature fusion \cite{sun2023enhancing, wang2023low}, multi-frame input for foreground-background separation and UAV classification \cite{craye2019spatio}, and adaptive frameworks combining appearance and motion features \cite{liu2025real}. While motion-aware techniques significantly improve detection performance, the computational overhead of optical flow estimation restricts their applicability in real-time scenarios.

\begin{figure*}[htbp]
	\centering
	\includegraphics[width=\linewidth]{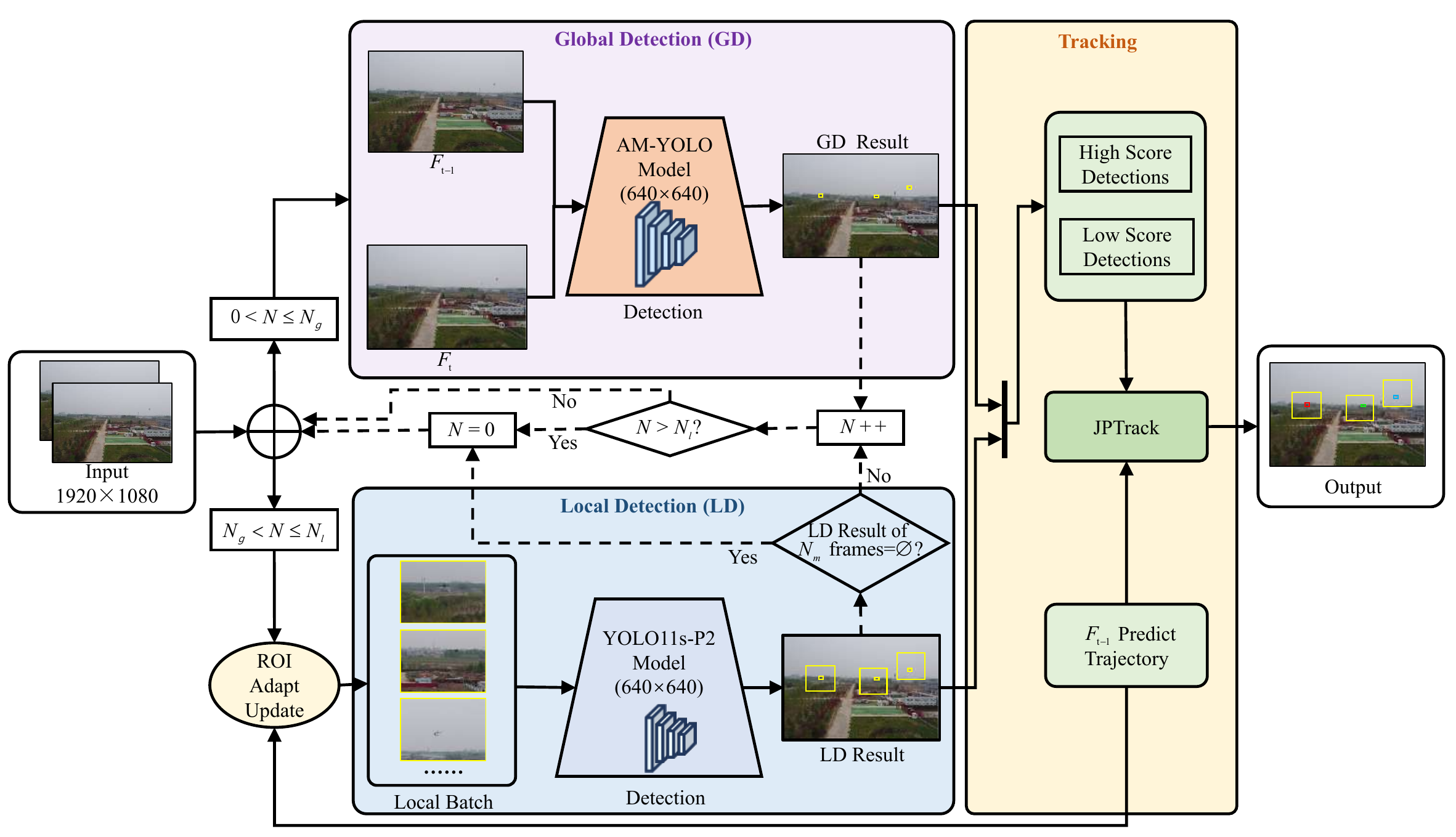}	
	\caption{The overall architecture of the proposed GL-DT framework, comprising global–local collaborative detection (GD and LD) and a Tracking module. GD utilizes AM-YOLO for the global detection, LD employs YOLO11s-P2 for local detection, and JPTrack ensures stable multi-object tracking.}
	\label{Fig2}	
\end{figure*}

\subsection{UAV Tracking Methods}
UAV tracking faces three fundamental challenges: high visual similarity among targets, small target size with weak feature representation, and complex motion patterns such as abrupt maneuvers, hovering, and diving--all of which significantly increase the difficulty of robust tracking. To address these issues, most existing studies adopt a tracking-by-detection paradigm, focusing on enhanced feature extraction, context modeling, and post-detection tracking optimization. For instance, Hong et al. \cite{hong2022real} leveraged 5G networks for cloud-based UAV video transmission and adopted an improved DeepSORT tracker, achieving real-time multi-target tracking. Ma et al. \cite{ma2022multi} integrated an optimized detector with DeepSORT to track multiple adapted UAVs within a 600-meter range; however, their method’s reliance on external appearance features increased the computational cost. Similarly, Delleji et al. \cite{delleji2022visual} enhanced DeepSORT with fused CNN features and a refined detector but required an auxiliary network for appearance embedding. Wang et al. \cite{wang2024target} introduced Transformer blocks into a Swin Transformer backbone and applied attention mechanisms to improve feature discrimination, achieving accurate UAV swarm tracking via ByteTrack’s data association—an effective strategy for dense small-object scenarios. Yuan et al. \cite{yuan2022learning} proposed an Adaptive Spatio-Temporal Context-Aware (ASTCA) model that learns contextual weights to differentiate targets from background and integrates spatial cues into a DCF tracker, improving robustness against scale variations and viewpoint shifts.

Some works adopt a joint detection and tracking framework. Zhao et al. \cite{zhao2022vision} developed the DUT Anti-UAV dataset and proposed a collaborative strategy integrating detectors and trackers, significantly boosting tracking accuracy. Cheng et al. \cite{cheng2022anti} presented a long-term tracking architecture (SiamAD) combining a Siamese network with re-detection modules, enhanced by hybrid attention and hierarchical discrimination, though it did not utilize motion cues. Li et al. \cite{li2023global} proposed an infrared-based anti-UAV framework that integrates global-local detection and tracking with motion and appearance features to handle occlusion and sparse visual information. However, its reliance on simultaneous global, local, and optical flow computations imposes high computational demands, limiting real-time applicability. While notable progress has been made in detection-tracking integration and feature enhancement, challenges remain in adapting to complex dynamic environments and balancing computational efficiency with real-time performance, highlighting key directions for future research.

\section{Methods}
We propose a real-time multi-object detection and tracking framework for UAVs, termed GL-DT (Global-Local Detection and Tracking), as illustrated in Fig. \ref{Fig2}. The framework follows the tracking-by-detection paradigm. integrates a global–local collaborative detection mechanism. By employing frame-level periodic scheduling, it dynamically switches between global and local modes, enabling complementary information exchange to balance large-scale situational awareness with fine-grained localization of small targets. On this basis, we further propose the JPTrack algorithm, designed to achieve efficient association and robust tracking of detection results.

Specifically, the overall framework comprises three core components: Global Detection (GD), Local Detection (LD), and Tracking. GD and LD are responsible for detection, while the Tracking module ensures target association across frames. The system takes two consecutive frames of 1920×1080 resolution as inputs. By jointly operating detection and tracking, it achieves robust UAV detection and stable tracking. 

\begin{figure*}[htbp]
	\centering
	\includegraphics[width=\linewidth]{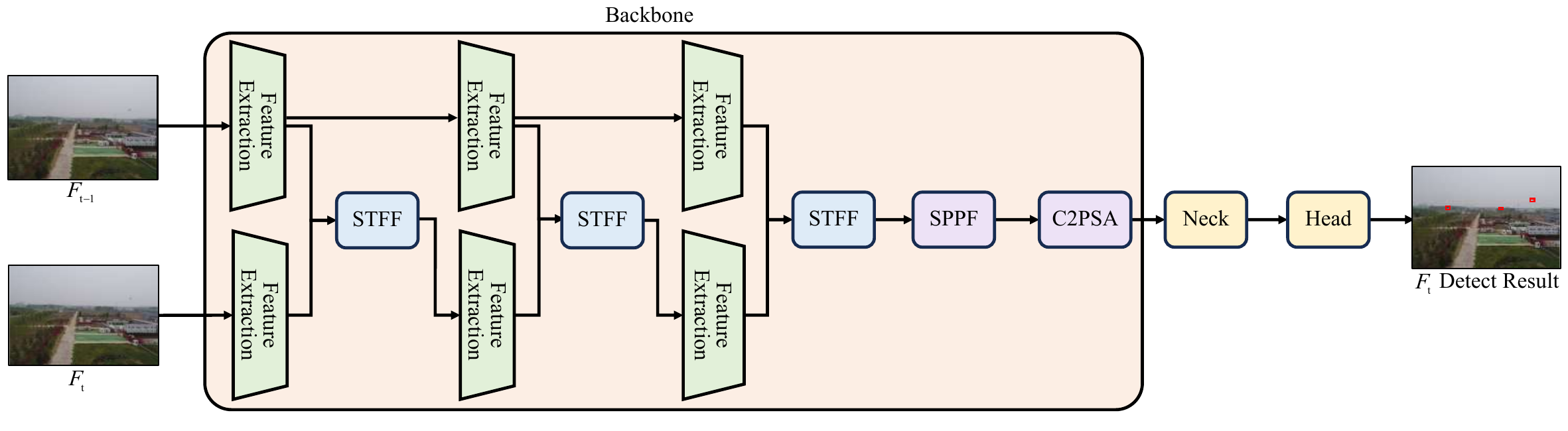}	
	\caption{AM-YOLO Model Architecture: Utilizes dual-frame input and incorporates a customized STFF module for multi-scale spatial-temporal feature extraction.}
	\label{Fig3}	
\end{figure*}

During the detection stage, the system adaptively switches between GD and LD modes according to the frame index $N$. By default, the system operates in GD mode to establish a comprehensive baseline for scene perception. In this mode, for frames satisfying $0 < N \leq N_g$, the system employs dual-frame input $F_t$ and $F_{t-1}$, and utilizes an improved AM-YOLO model (rescaled to $640 \times 640$ during inference) to enhance spatio-temporal feature modeling. Once stable tracking of the target is achieved in GD mode, the system automatically switches to LD mode. For frames satisfying $N_g < N \leq N_l$, the system generates Regions of Interest (ROIs) based on the previously confirmed target locations. The YOLO11s-P2 model then performs parallel inference on these candidate ROIs (also rescaled to $640 \times 640$), thereby enabling fast and stable detection of the filtered targets. If the LD mode persists for more than $N_l$ frames, the system automatically reverts to GD mode to ensure timely capture of newly appearing targets or those outside the local perceptual range, effectively reducing the risk of missed detections. Moreover, to address potential target loss during LD mode, a reset mechanism is embedded. If no target is detected within any ROI for $N_m$ consecutive frames, the system considers the current local context invalid and immediately resets to GD mode to reacquire global scene information. During tracking, the proposed JPTrack algorithm performs hierarchical association and matching of high- and low-confidence detection results, thereby ensuring stable target tracking across consecutive frames. In summary, the GL-DT framework integrates an adaptive global-local collaborative detection mechanism with the JPTrack tracking algorithm, which, while maintaining real-time performance, significantly enhances the capability of the system to achieve continuous detection and stable tracking of small UAV targets under complex and challenging scenarios.

\subsection{Global Detection}
In the GL-DT framework, the GD module is built upon the YOLO11 architecture, and an enhanced detection model, named AM-YOLO, is proposed. This model integrates a multi-scale spatiotemporal feature modeling mechanism, significantly enhancing its capability to detect dynamic objects. The overall structure of the model is illustrated in Fig. \ref{Fig3}.

The AM-YOLO model takes as input the current frame ${F_t}$ and the previous frame ${F_{t-1}}$,  processing them through a weight-shared backbone network to extract multi-layer semantic features. To effectively capture dynamic variations across both spatial and temporal dimensions, several critical stages of the backbone incorporate an STFF module, enabling deep cross-frame feature collaboration. The fused features are subsequently refined using a Spatial Pyramid Pooling Fast (SPPF) module and a Cross Stage Partial with Pyramid Squeeze  Attention (C2PSA) mechanism to enhance receptive field adaptation and contextual representation. These enriched features are then aggregated in the Neck module and passed to the Head for precise object detection outputs.
 
\subsubsection{STFF module}
\begin{figure*}[htbp]
	\centering
	\includegraphics[width=\linewidth]{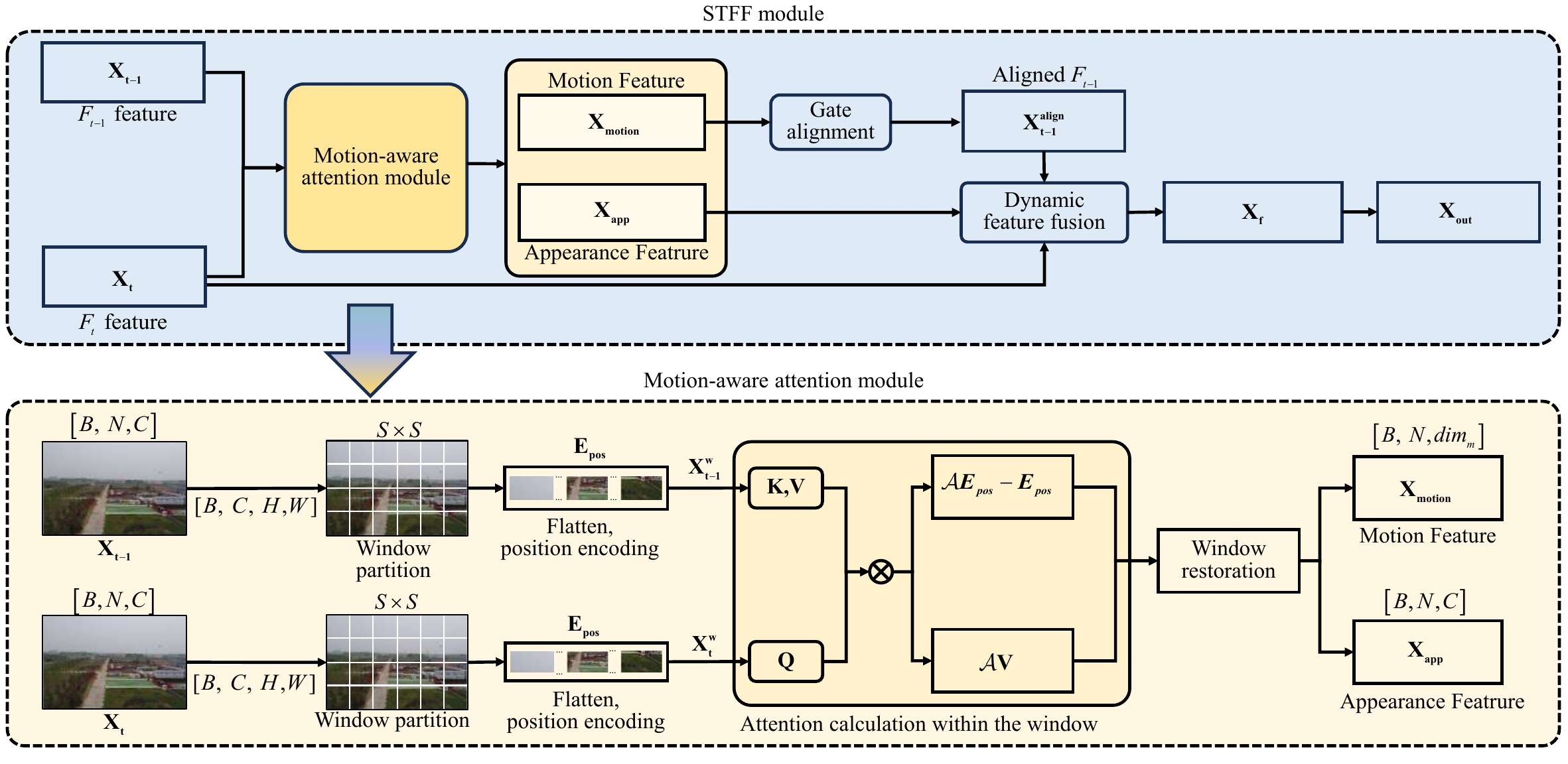}	
	\caption{Framework of the STFF module and Motion-aware attention module. STFF integrates spatio-temporal features via Motion-aware attention and feature fusion, while the attention module extracts appearance and motion features through local window operations.}
	\label{Fig4}	
\end{figure*}

As shown in Fig. \ref{Fig4}, the STFF module is made to effectively integrate spatio-temporal characteristics across neighboring frames, improving the model's capacity to capture dynamic object patterns and preserve temporal consistency.  The main concept is the introduction of a Motion-aware Attention Module that allows for deep fusion and dynamic alignment of data from successive frames, accurately mimicking object motion characteristics and temporal fluctuations.

Specifically, the STFF module takes the current frame feature map $\mathbf{X_t}$ and the previous frame feature map $\mathbf{X_{t-1}}$ as inputs. It first employs the Motion-aware Attention Module to extract two critical types of features: the appearance feature $\mathbf{X_{app}}$, which preserves semantic discriminability, and the motion feature $\mathbf{X_{motion}}$, which models the temporal displacement of objects. Subsequently, based on the motion feature, the module applies a gated alignment mechanism to achieve feature alignment, where the motion feature is transformed into gating weights $\mathbf{G}$ through a $1\times1$ convolution followed by a Sigmoid activation function:
\begin{equation}
\mathbf{G} =\text{Sigmod}(\text{Conv}_{1\times1}(\mathbf{X_{motion}}))
\label{eq1}
\end{equation}

This gating weight acts on the features of the previous frame to achieve adaptive alignment:
\begin{equation}
\mathbf{X_{t-1}^{align}}= \mathbf{{X}_{t-1}} \odot \mathbf{G}
\label{eq2}
\end{equation}
where $\odot$ denotes element-wise multiplication. This gating mechanism adaptively adjusts the feature response strength according to motion information, thereby enabling efficient feature alignment.

Further, $\mathbf{X_{t-1}^{\text{align}}}$, $\mathbf{X_{app}}$, and $\mathbf{X_t}$ are jointly fed into the dynamic feature fusion module, where convolution and Softmax operations are employed to adaptively generate the fusion weights:
\begin{equation}
\mathbf{{[W_{1},W_{2},W_{3}]}} =\text{Softmax}(\text{Conv}_{1\times1}([\mathbf{X_t}; \mathbf{X_{app}}; \mathbf{X_{t-1}^{align}]}))
\label{eq3}
\end{equation}

Then the fusion feature $\mathbf{{X}_{f}}$ is obtained through weighted summation, as
\begin{equation}
\mathbf{X_f} = \mathbf{W_1} \odot \mathbf{X_t} + \mathbf{W_2} \odot \mathbf{X_{app}} + \mathbf{W_3} \odot \mathbf{X_{t-1}^{align}}
\label{eq4}
\end{equation}

The final output feature $\mathbf{X_{out}}$  adopts a residual fusion mechanism, retaining the original features while introducing temporal information:
\begin{equation}
\mathbf{X_{out}} = \alpha \cdot \mathbf{X_f} + (1 - \alpha) \cdot \mathbf{X_t}
\label{eq5}
\end{equation}
among them, $\alpha$ is the learnable weighting coefficient.

\subsubsection{Motion-aware attention module}
To overcome the limitations of traditional optical flow methods in semantic modeling and avoid the high computational complexity caused by global self-attention, this paper proposes a Motion-aware attention module based on local windows, which jointly captures spatial and temporal dependencies across consecutive frames. The module partitions the feature maps into non-overlapping local windows, constructs local sequences for the current and preceding frames within each window, and incorporates motion-sensitive positional encoding to enhance spatial awareness, as illustrated in Fig. \ref{Fig4}.

Specifically, the input features from the current frame $\mathbf{X_t}$ and the previous frame $\mathbf{X_{t-1}}$ both have a shape of $[B, M, C]$, where $B$ denotes the batch size, $M = H \times W$ represents the number of spatial locations, $C$ is the number of channels, $H$ is the height of the feature map, and $W$ is its width. The input features are first rearranged from $[B, M, C]$ into $[B, C, H, W]$, after which the two-dimensional feature maps are partitioned into non-overlapping local windows of size $S \times S$. To enhance spatial awareness, the module introduces normalized two-dimensional coordinates and employs a linear projection to generate the action-sensitive positional encoding $\mathbf{E_{pos}}$. Within each window, the query vectors are derived from the current frame window $\mathbf{X_t^w}$, while the key–value pairs $[\mathbf{K}, \mathbf{V}]$ are obtained from the previous frame window $\mathbf{X_{t-1}^w}$:
\begin{equation}
\mathbf{Q} = \mathbf{W_q} \mathbf{X_t^w}, \quad [\mathbf{K}, \mathbf{V}] = \mathbf{W_{kv}} \mathbf{X_{t-1}^w}
\label{eq6}
\end{equation}
where $\mathbf{W_q}$ and $\mathbf{W_{kv}}$ are the respective linear projection matrices for queries and key-value pairs.

Attention weights  $\mathcal{A}$ are computed using the scaled dot-product attention mechanism:
\begin{equation}
\mathcal{A} = \text{Softmax} \left( \frac{\mathbf{Q} \mathbf{K}^T}{\sqrt{d_k}} \right), d_k=C/N_{heads}
\label{eq7}
\end{equation}
here, $N_{heads}$ is the number of attention heads. The scaling factor $\frac{1}{\sqrt{d_k}}$ stabilizes gradient propagation during training. 

Finally, the outputs from all local windows are reassembled to recover the original spatial structure, yielding two separate feature representations: $\mathbf{{X}_{app}}$ and $\mathbf{{X}_{motion}}$, as
\begin{equation}
\mathbf{X_{app}} = \mathcal{A} \mathbf{V}, \quad \mathbf{X_{motion}} = \mathcal{A} \mathbf{E_{pos}} - \mathbf{E_{pos}}
\label{eq8}
\end{equation}
where $\mathbf{X_{app}}$ is computed by applying the attention weights to the value vectors $\mathbf{V}$, resulting in a tensor of shape $[B, N, C]$. The motion feature $\mathbf{X_{motion}}$ is derived from the difference between the attention-weighted positional encoding and the original positional encoding, resulting in a tensor of shape $[B, N, dim_m]$, where $dim_m$ is the dimensionality of the motion feature. After window restoration, $\mathbf{X_{app}}$ is reshaped back to $[B, C, H, W]$, while $\mathbf{X_{motion}}$ is processed accordingly for subsequent fusion steps.

This design not only effectively models the motion changes within the local area but also enhances the motion perception ability through explicit position coding, providing more discriminative spatio-temporal feature representations.

\subsection{Local Detection}
To enhance detection accuracy and robustness, we introduce an LD module that performs fine-grained identification on candidate target regions filtered by the global stage. This approach effectively suppresses background interference and improves target discriminability. The performance of the LD module relies on the accurate extraction of ROIs, which is adaptively updated based on trajectory predictions from the previous frame to ensure continuous focus on potential target areas. Following the modeling approach with Kalman filtering described in \cite{liu2025real}, the system predicts the trajectory point in the current frame, and a $300\times300$ detection window centered at this point is constructed as the local detection region.

\begin{figure*}[htbp]
	\centering
	\includegraphics[width=\linewidth]{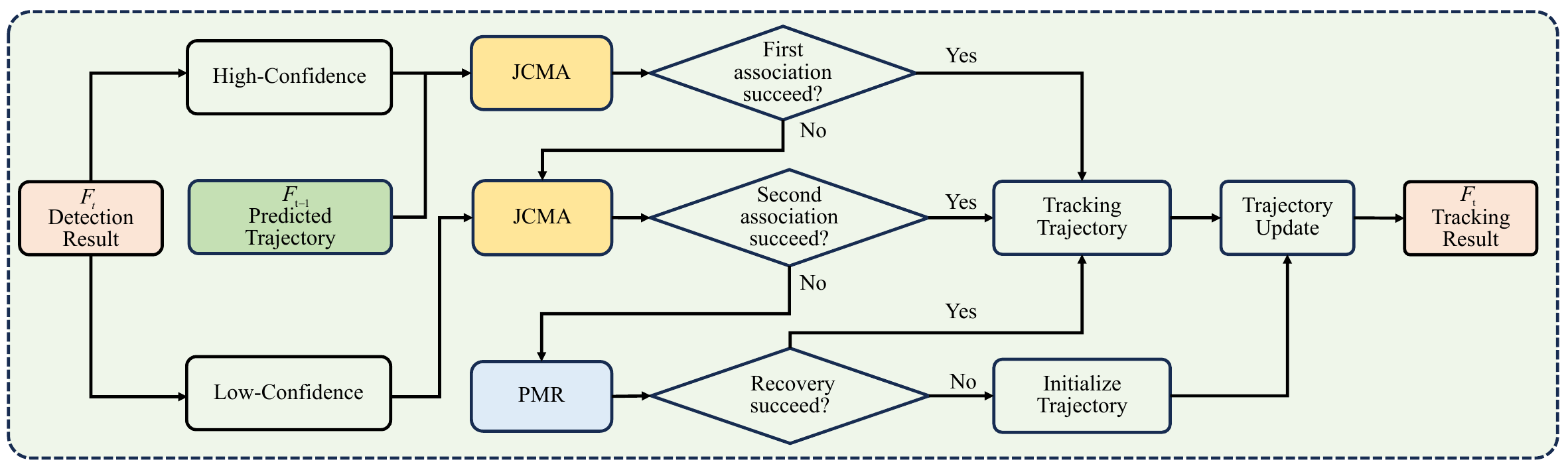}	
	\caption{JPTrack Tracking Flowchart: Composed of JCMA and PMR Modules. The JCMA constructs a comprehensive matching cost by fusing multi-dimensional features including IoU, distance, motion consistency, and geometric relationships; the PMR module probabilistically models the historical states of lost trajectories using Gaussian Mixture Models to achieve trajectory recovery after short-term occlusions.}
	\label{Fig5}	
\end{figure*}

Given the dynamic movement of targets across consecutive frames, ROIs may spatially overlap or contain multiple targets that are significantly distant from each other. To enhance detection efficiency and regional adaptability, we propose an adaptive ROI update mechanism that dynamically analyzes the spatial relationships between ROIs and targets. Specifically, when the overlap between two ROIs exceeds the threshold $\tau_o$, they are merged to eliminate redundant computation. If multiple targets are present within a single ROI and their spatial separation increases over time, the ROI is first adaptively expanded to ensure full coverage. Once the distance between targets surpasses the threshold $\tau_d$ pixels, the original ROI is segmented into smaller regions to improve detection granularity and precision. Throughout this process, the ROI’s position and size are continuously adjusted to ensure that all targets remain within a predefined ‘safe zone’ $\tau_s$, defined as the central region covering $\tau_s$ of the ROI’s total area, thereby maintaining reliable localization and stable detection performance.

For implementation, to balance detection accuracy with real-time processing requirements, we adopt the YOLO11s-P2 model, which offers strong performance in small object detection with relatively low computational complexity. Considering that multiple targets may be distributed across multiple ROIs in practical scenarios, conventional sequential ROI processing presents efficiency bottlenecks. To address this, we batch all ROIs within the same frame into a single input for the model, enabling parallel processing of Local Detection. This strategy significantly improves inference throughput while maintaining detection accuracy, thereby meeting the real-time demands of complex and dynamic environments.

\subsection{Tracking}
The tracking process is handled by JPTrack, a robust multi-object tracker designed to address the challenges of irregular motion and frequent occlusions in UAV scenarios. While built upon the multi-stage matching framework of ByteTrack, JPTrack introduces two core innovations—the Joint Cost Matching Association (JCMA) module and the Probability-driven Memory Recovery (PMR) module—which collectively enhance tracking continuity and mitigate identity switches, as illustrated in Fig. \ref{Fig5}.

JPTrack adopts a three-stage matching strategy to maximize the association accuracy between detections and trajectories. First, the detections in the current frame are categorized into high-confidence and low-confidence groups based on the confidence thresholds $t_h$ and $t_l$, which correspond to the high and low confidence levels, respectively. In the first stage, the JCMA module prioritizes matching high-confidence detections with all active trajectories; successfully matched trajectories are updated immediately, while unmatched items proceed to the second stage. In the second stage, remaining unmatched trajectories are matched with low-confidence detections via JCMA, and successfully matched trajectories are updated; unmatched targets advance to the third stage. In the third stage, the PMR module leverages Gaussian Mixture Model (GMM) modeling of historical trajectories for backtracking. Successfully matched detections are restored as valid trajectories; if matching fails, high-confidence detections initialize new trajectories, while low-confidence detections are discarded. This staged process, augmented by JCMA's robust association and PMR's memory-driven recovery, forms the distinctive core of the JPTrack framework.

\subsubsection{JCMA}
To overcome the limitations of IoU-only matching under rapid or irregular motion, the JCMA module computes a composite cost matrix by integrating multiple motion and spatial cues. The matching cost between the $i$-th predicted trajectory $T_i$ and the $j$-th detection $D_j$ is defined as:  
\begin{equation}
\text{Cost}_{i,j} = \omega_1 \cdot C_{i,j}^{\text{iou}} + \omega_2 \cdot C_{i,j}^{\text{dist}} + \omega_3 \cdot C_{i,j}^{\text{motion}} + \omega_4 \cdot C_{i,j}^{\text{rel}}
\label{eq9}
\end{equation}
where $C_{i,j}^{\text{iou}}$ denotes the IoU overlap cost, $C_{i,j}^{\text{dist}}$ the normalized distance cost, $C_{i,j}^{\text{motion}}$ the motion consistency cost, and $C_{i,j}^{\text{rel}}$ the geometric relational cost. The weighting coefficients $\omega_1, \omega_2, \omega_3, \omega_4$ are optimized empirically to balance the contribution of each term, as detailed in Table~\ref{tab1}. 

The specific calculations of each cost item are as follows. $C_{i,j}^{iou}$ is used to measure the geometric overlap between the $i$-th predicted trajectory $T_i$ and the $j$-th detection box $D_j$:
\begin{equation}
C_{i,j}^{\text{iou}} = 1 - \text{IoU}(T_i, D_j)
\label{eq10}
\end{equation}

$C_{i,j}^{dist}$ accounts for the Euclidean distance cost between the center of the predicted trajectory and that of the detection box:
\begin{equation}
C_{i,j}^{\text{dist}} = \min\left(1.0, \frac{\|\mathbf{c}i - \mathbf{c}_j\|_2}{s_{i,j}}\right)
\label{eq11}
\end{equation}
where $\mathbf{c}_i$ and $\mathbf{c}_j$ denote the center coordinates of trajectory $T_i$ and detection $D_j$, respectively. The adaptive scale factor $s_{i,j}$ is defined as:  
\begin{equation}
s_{i,j} = \frac{w_i + h_i + w_j + h_j}{4} \times 2.0
\label{eq12}
\end{equation}

$C_{i,j}^{\text{motion}}$ incorporates speed, direction, and acceleration,  and is defined as follows:
\begin{equation}
C_{i,j}^{\text{motion}} = \beta_1 \cdot C_{\text{speed}} + \beta_2 \cdot C_{\text{direction}} + \beta_3 \cdot C_{\text{acceleration}}
\label{eq13}
\end{equation}
with  
\begin{equation}
\begin{cases}
C_{\text{speed}} = \min\left(1.0, \frac{|v_{\text{expected}} - v_{\text{avg}}|}{\max(50.0, 2.0 + v_{\text{avg}})}\right) \\
C_{\text{direction}} = \min\left(1.0, \frac{|\Delta\theta|}{\pi/2}\right) \\
C_{\text{acceleration}} = \min\left(1.0, \frac{\|\mathbf{a}_{\text{expected}}\|_2}{30.0}\right)
\end{cases}
\label{eq14}
\end{equation}
where $v_{\text{expected}} = \|\mathbf{c}_j - \mathbf{c}_i\|_2$ is the expected velocity, $v_{\text{avg}}$ is the historical average velocity, $\Delta\theta$ is the angular difference between expected and historical directions, and $\mathbf{a}_{\text{expected}}$ denotes the expected acceleration vector.  

$C_{i,j}^{\text{rel}}$ measures the structural similarity between the predicted trajectory and the candidate detection set:  
\begin{equation}
C_{i,j}^{\text{rel}} = 1 - \frac{1}{M}\sum_{k=1}^{M} \frac{(\mathbf{c}_k - \mathbf{c}_i) \cdot (\mathbf{c}_k - \mathbf{c}_j)}{\|\mathbf{c}_k - \mathbf{c}_i\|_2 \cdot \|\mathbf{c}_k - \mathbf{c}_j\|_2}
\label{eq15}
\end{equation}
where $\mathbf{c}_k$ denotes the center coordinates of the $k$-th object (excluding $T_i$ and $D_j$), and $M$ is the total number of such objects.  

By incorporating trajectory state-awareness and global spatial relationships, this joint cost matrix matching strategy significantly mitigates identity switches and matching errors arising from motion uncertainties.

\subsubsection{PMR}
The PMR module is designed to recover trajectories after short-term occlusions or detection failures.  For each lost trajectory, it constructs a probabilistic motion model using a GMM to capture the multi-modal uncertainty in the target's historical states. For each lost track $T_i$, an 8-dimensional feature vector $\mathbf{Y}_i$ is extracted from its historical states:
\begin{equation}
    \mathbf{Y}_i = \left[ x_i^{\text{abs}}, y_i^{\text{abs}}, x_i^{\text{rel}}, y_i^{\text{rel}}, v_{x,i}, v_{y,i}, \theta_i, w_i \right]^T
    \label{eq16}
\end{equation}
where $(x_i^{\text{abs}}, y_i^{\text{abs}})$ denotes the global position; $(x_i^{\text{rel}}, y_i^{\text{rel}})$ is the relative position within the ROI; $(v_{x,i}, v_{y,i})$ are the velocity components; $\theta_i$ represents the motion direction; $w_i$ is the time-decay weight factor, computed as: 
\begin{equation}
    w_i = \exp(-\gamma \cdot (t_{\text{current}} - t_i))
    \label{eq17}
\end{equation}
where $\gamma=0.1$ is the attenuation coefficient, $t_{\text{current}}$ represents the timestamp of the current frame, and $t_i$ is the timestamp of the frame where the sample is located.

To adapt to varying motion complexities in UAV applications, the number of GMM components $K_{\text{adaptive}}$ is dynamically determined:
\begin{equation}
    K_{\text{adaptive}} = \min\left(2, \max\left(1, \left\lfloor \frac{N_{\text{samples}}}{3} \right\rfloor\right)\right)
    \label{eq18}
\end{equation}
where $N_{samples}$ represents the number of historical samples.

Given the set of historical feature vectors $\{\mathbf{Y}_1, \mathbf{Y}_2, ..., \mathbf{Y}_n\}$  for a lost track $T_i$, the GMM models the probability density function for a single feature vector $\mathbf{Y_i}$:
\begin{equation}
p(\mathbf{Y_i}|\Theta) = \sum_{k=1}^{K} \pi_k \mathcal{N}(\mathbf{Y_i}|\boldsymbol{\mu}_k, \boldsymbol{\Sigma}_k)
\label{eq19}
\end{equation}
where $\Theta = \{\pi_k, \boldsymbol{\mu}_k, \boldsymbol{\Sigma}k\}_{k=1}^{K}$ includes the mixture weights $\pi_k$, means $\boldsymbol{\mu}_k$, and covariances $\boldsymbol{\Sigma}_k$.

Then, the model parameters are optimized through the EM algorithm \cite{bishop2006pattern} to model the historical features. For a candidate detection $D_j$ in the current frame, we first extract its feature vector $\mathbf{Z}_j$ using the same feature representation as defined in Equation~\ref{eq16}, and its matching probability with the lost trajectory is computed using the GMM:
\begin{equation}
P_{\text{match}}(D_j|T_i) = \sum_{k=1}^{K} \pi_k \mathcal{N}(\mathbf{Z}_j|\boldsymbol{\mu}_k, \boldsymbol{\Sigma}_k)
\label{eq20}
\end{equation}

To ensure the accuracy of recovery, the system imposes a constraint based on time delay:
\begin{equation}
C_{\text{time}} = \frac{1}{N} \sum_{i=1}^{N} w_i
\label{eq21}
\end{equation}

The final matching score is defined as:
\begin{equation}
P_{\text{final}} = P_{\text{match}} \cdot C_{\text{time}}
\label{eq22}
\end{equation}

If $P_{\text{final}} > \tau_t$, the system considers $D_j$ to be a successful match for trajectory $T_i$, enabling its recovery. The threshold  $\tau_t$ was chosen empirically to balance recovery rate and matching accuracy. By integrating the JCMA module for precise detection-to-track matching and the PMR module for memory-guided recovery, this tracking framework effectively addresses the challenges of identity switches and trajectory fragmentation commonly encountered in complex UAV scenarios. Experimental results demonstrate that the proposed method significantly enhances the continuity and stability of MOT while maintaining real-time performance.

\section{Experiment}
\subsection{Dataset}

In evaluating the performance of our proposed detection method, we selected two challenging video datasets to ensure a comprehensive and accurate assessment.

1) MOT-FLY Dataset:
The MOT-FLY dataset \cite{zhaochen2024vision} is collected for multi-UAV target tracking in air-to-air scenarios. It comprises 16 RGB video sequences, from which a total of 11,186 frames have been extracted. Among these, 7,238 frames are used for training and 3,948 for testing, all at a uniform resolution of 1920×1080. Each video contains between one and three UAV targets, with diverse scene configurations that encompass viewpoint variations, complex backgrounds, illumination changes, and various motion patterns, collectively enhancing the task’s complexity. Notably, the majority of target instances occupy less than 5\% of the image area, significantly increasing the challenge of both detection and tracking.

\begin{figure}[htbp]
	\centering
	\includegraphics[width=\linewidth]{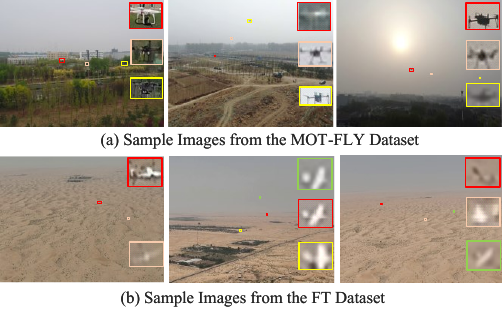}	
	\caption{Sample images from the MOT-FLY and FT Datasets.}
	\label{Fig6}	
\end{figure}

2) FT Dataset:
The FT Dataset is a custom-built collection specifically designed for visual detection and multi-object tracking of fixed-wing UAVs in complex environments. The dataset is collected using a DJI H20T series camera mounted on a platform in desert conditions and comprises 19 high-definition video sequences at a resolution of 1920×1080. Each sequence contains one to three small-scale fixed-wing UAVs captured at long range, with a total of 25,855 frames, with an average pixel occupancy below 0.1\%. Due to the extremely low pixel ratio of the targets, both detection and tracking tasks present significant challenges. For systematic performance evaluation, we partitioned the first 60\% of frames in each sequence as the training set and the remaining 40\% as the test set, enabling a comprehensive assessment of model performance in multi-object detection and tracking. Fig. \ref{Fig6} presents representative image samples from the MOT-FLY and FT datasets.

\subsection{Evaluation Metrics and Implementation Details}

\subsubsection{Evaluation Metrics}
To comprehensively evaluate the performance of the proposed algorithm, this study employs the following widely adopted metrics: ID Switch (IDSW), Identification F-Score (IDF1), Multiple Object Tracking Accuracy (MOTA), Multiple Object Tracking Precision (MOTP), Higher Order Tracking Accuracy (HOTA), Detection Accuracy (DetA), Association Accuracy (AssA), and Frames Per Second (FPS). These metrics provide a multi-dimensional assessment of the model, systematically and thoroughly evaluating identity consistency, tracking accuracy, and algorithmic real-time performance. Specifically:

IDF1: This score evaluates the consistency of object identities, representing the proportion of correctly identified detections. It is the harmonic mean of Identification Precision (IDP) and Identification Recall (IDR):
\begin{equation}
\mathrm{IDF1} = 2 \times \frac{\mathrm{IDP} \times \mathrm{IDR}}{\mathrm{IDP} + \mathrm{IDR}}
\label{eq23}
\end{equation}

This metric emphasizes temporal identity continuity and re-identification ability; higher values indicate a stronger capability of maintaining consistent object identities.

MOTA: Measuring the overall performance of a multi-object tracking system by accounting for false positives (FP), false negatives (FN), and IDSW. It is defined:
\begin{equation}
\mathrm{MOTA} = 1 - \frac{\sum (\mathrm{FP} + \mathrm{FN} + \mathrm{IDSW})}{\sum \mathrm{GT}}
\label{eq24}
\end{equation}
where $\mathrm{GT}$ denotes the total number of ground-truth objects, a MOTA value closer to 1 indicates better overall tracking performance.

MOTP: measuring the average overlap between predicted and ground-truth object positions. It is calculated as:
\begin{equation}
\mathrm{MOTP} = \frac{\sum_{i,t} d_{i,t}}{\sum_t c_t}
\label{eq25}
\end{equation}
where  $d_{i,t}$ denotes the IoU distance between matched predicted and ground-truth objects at time $t$, and $c_t$ is the number of successfully matched objects in frame; higher MOTP values indicate more precise object localization by the tracker.

HOTA: provides a unified evaluation of detection and association performance, balancing object detection accuracy and trajectory association quality. It is defined as
\begin{equation}
\mathrm{HOTA}_\alpha = \sqrt{\frac{\sum_{c \in \{\text{TP}\}} A(c)}{|\text{TP}| + |\text{FN}| + |\text{FP}|}}
\label{eq26}
\end{equation}
where $c$ denotes a correctly matched sample, and $\text{TP}$, $\text{FN}$, and $\text{FP}$ represent the numbers of true positives, false negatives, and false positives at the detection level, respectively. The local association accuracy $A(c)$ is defined as:
\begin{equation}
A(c) = \frac{|\text{TPA}(c)|}{|\text{TPA}(c)| + |\text{FNA}(c)| + |\text{FPA}(c)|}
\label{eq27}
\end{equation}
Here, $\text{TPA}(c)$ refers to the correctly associated pairs where both the predicted ID and ground-truth ID correspond to $c$; $\text{FNA}(c)$ denotes cases where the ground-truth ID is $c$ but the predicted ID is not $c$, or when $c$ appears in the $\text{FN}$ set; $\text{FPA}(c)$ denotes cases where the predicted ID is $c$ but the ground-truth ID is not $c$, or when $c$ appears in the $\text{FP}$ set.

DetA: A metric measuring detection accuracy, defined as:
\begin{equation}
\mathrm{DetA} = \frac{|\text{TP}|}{|\text{TP}| + |\text{FN}| + |\text{FP}|}
\label{eq28}
\end{equation}

AssA: A metric used to evaluate the ability of a model to maintain identity consistency across trajectories, defined as:
\begin{equation}
\mathrm{AssA}_\alpha = \frac{\sum_{c \in \text{TP}} A(c)}{|\text{TP}|}
\label{eq29}
\end{equation}

\subsubsection{Implementation Details}
The experiments in this study were conducted on a high-performance computing platform equipped with dual NVIDIA GeForce RTX 3090 GPUs. During the training phase, an enhanced YOLO11 architecture was employed, with all input images uniformly resized to 640×640 pixels. The optimizer used was Stochastic Gradient Descent (SGD) with a momentum coefficient of 0.937 and an initial learning rate set to 0.01. Training was performed for a total of 200 epochs with a batch size of 32. To improve the model’s generalization and adaptability to the specific task, fine-tuning was conducted using a pre-trained model on UAV-related datasets.

The GL-DT system comprises a Global Detection model and a Local Detection model, each requiring separate training. The Global Detection model is trained on full images using a custom-designed AM-YOLO architecture, while the Local Detection model is trained on target regions cropped from the original images using the YOLO11s-P2 architecture. Upon completion of training, the two models are integrated during the inference stage to enhance UAV detection and tracking performance. The system involves the configuration of several key parameter thresholds, all of which have been systematically validated through extensive ablation experiments to ensure an optimal balance between accuracy and reliability, as summarized in Table~\ref{tab1}.

\begin{table}[htbp]
    \centering
    \caption{Threshold Parameter Configuration}
    \begin{tabular}{c|c||c|c||c|c}
        \hline
        \textbf{Notation} & \textbf{Value} & \textbf{Notation} & \textbf{Value} & \textbf{Notation} & \textbf{Value} \\
        \hline
        $N_g$       & 30     & $N_l$      & 120     & $\alpha$      & 0.1 \\
        $\tau_o$    & 0.2    & $\tau_d$   & 700     & $\tau_s$      & 0.8 \\
        $\omega_1$  & 0.3    & $\omega_2$ & 0.3     & $\omega_3$    & 0.2 \\
        $\omega_4$  & 0.2    & $\beta_1$ & 0.4     & $\beta_2$    & 0.4 \\
        $\beta_3$  & 0.2    & $\gamma $  & 0.1     & $\tau_t$      & 0.6 \\
        \hline
    \end{tabular}
    \label{tab1}
\end{table}

\subsection{Comparison with Prior Works}

To comprehensively validate the performance advantages of the proposed multi-object tracking method GL-DT, we conducted comparative experiments on the MOT-FLY and FT datasets and systematically evaluated it against a range of state-of-the-art algorithms, as summarized in Table~\ref{tab2}. Specifically, ByteTrack \cite{zhang2022bytetrack}, BoT-SORT \cite{aharon2022bot}, DeepSORT \cite{wojke2017simple}, and OC-SORT \cite{cao2023observation} represent the classical tracking-by-detection paradigm, for which we consistently employed the YOLO11s-P2 model as the detector to ensure input uniformity. In contrast, FairMOT \cite{zhang2021fairmot}, CenterTrack \cite{zhou2020tracking}, TrackerFormer \cite{meinhardt2022trackformer}, and MOTR \cite{zeng2022motr} exemplify the joint detection and tracking paradigm, all implemented based on their official open-source releases. This comparison encompasses the major contemporary approaches in multi-object tracking, providing a rigorous assessment of the accuracy and robustness of our method.

\begin{table*}[htbp]
  \centering
  \caption{Comparison of Tracking Methods}
  \begin{tabular}{c|c|c|c|c|c|c|c|c|c}
    \hline
    \textbf{Dataset} & Model & IDSW$\downarrow$ & IDF1$\uparrow$ & MOTA$\uparrow$ & MOTP$\uparrow$ & HOTA$\uparrow$ & DetA$\uparrow$ & AssA$\uparrow$ & FPS$\uparrow$ \\
    \hline
    \multirow{9}{*}{\centering MOT-FLY Dataset} 
    & ByteTrack \cite{zhang2022bytetrack}    & 122 & 52.42 & 50.84 & 80.41 & 39.20 & 41.99 & 36.74 & 35.81 \\
    & BoT-SORT \cite{aharon2022bot}     & 112 & 52.03 & 51.18 & 80.76 & 40.08 & 42.45 & 37.94 & 27.69 \\
    & DeepSORT \cite{wojke2017simple}      & 120 & 47.28 & 34.92 & 61.91 & 31.89 & 33.57 & 30.76 & 38.95 \\
    & OC-SORT \cite{cao2023observation}        & 77  & 49.76 & 48.26 & 80.87 & 36.75 & 39.06 & 34.26 & \textbf{43.06} \\
    & CenterTrack \cite{zhou2020tracking}   & 144 & 55.68 & 67.82 & 77.17 & 49.23 & 59.39 & 41.23 & 34.12 \\
    & FairMOT \cite{zhang2021fairmot}       & 138 & 58.96 & 69.41 & 79.65 & 48.08 & 55.01 & 43.53 & 36.88 \\
    & TrackFormer \cite{meinhardt2022trackformer}  & 106 & 62.71 & 74.61 & 77.85 & 54.21 & 61.74 & 48.18 & 6.52 \\
    & MOTR \cite{zeng2022motr}        & 86  & 68.76 & 76.48 & 78.64 & 56.51 & 63.22 & 50.76 & 12.26 \\
    & GL-DT (Ours)  & \textbf{33}  & \textbf{79.21} & \textbf{82.84} & \textbf{82.11} & \textbf{65.22} & \textbf{67.50} & \textbf{63.08} & 35.06 \\
    \hline
    \multirow{9}{*}{\centering FT Dataset} 
    & ByteTrack     & 76  & 46.03 & 66.87 & 75.58 & 43.02 & 52.89 & 38.03 & 36.43 \\
    & BoT-SORT       & 60  & 48.66 & 69.00 & 68.23 & \textbf{62.69} & 55.56 & 53.48 & 25.58 \\
    & DeepSORT      & 61  & 30.17 & 29.34 & 65.37 & 29.87 & 41.72 & 23.42 & 39.54 \\
    & OC-SORT       & 42  & 47.53 & 64.14 & \textbf{78.56} & 44.10 & 51.34 & 41.85 & \textbf{43.14} \\
    & CenterTrack   & 108 & 43.00 & 64.95 & 75.13 & 48.45 & 51.22 & 44.32 & 33.62  \\
    & FairMOT       & 92  & 32.50 & 63.01 & 74.99 & 46.64 & 54.70 & 42.75 & 37.81  \\
    & TrackFormer   & 72  & 68.75 & 68.42 & 74.82 & 52.42 & 56.04 & 48.72 & 6.80   \\
    & MOTR          & 68  & 70.96 & 69.42 & 73.51 & 53.48 & 56.43 & 50.72 & 11.72  \\
    & GL-DT (Ours)  & \textbf{14}  & \textbf{83.99} & \textbf{72.06} & 74.45 & 60.73 & \textbf{56.89} & \textbf{65.88} & 35.88 \\
    \hline
  \end{tabular}
  \label{tab2}
\end{table*}

As shown in Table~\ref{tab2}, the proposed GL-DT method demonstrates a clear and consistent advantage across both datasets. On the MOT-FLY dataset, GL-DT achieves the lowest number of IDSW (33), which is not only substantially lower than the maximum value of 77 from tracking-by-detection methods but also outperforms the best result (86) from joint detection and tracking approaches. This highlights its remarkable capability in maintaining identity consistency. In addition, GL-DT attains an IDF1 score of 79.21, surpassing the second-best method, MOTR, by 6.36\%, further underscoring its superior performance in identity consistency and trajectory matching accuracy. From the perspective of overall metrics, GL-DT also achieves state-of-the-art results in MOTA (82.84), MOTP (82.11), and HOTA (65.22), fully demonstrating its dual optimization capability in both identity preservation and localization accuracy for multi-object tracking. On the FT dataset, GL-DT exhibits the same level of robustness. It reduces IDSW to as low as 14 and achieves an IDF1 of 83.99, significantly outperforming all competing methods, which confirms its ability to maintain excellent identity consistency and trajectory continuity even in challenging scenarios. Moreover, GL-DT secures the top performance in MOTA (72.06) and achieves the highest score in AssA (65.88), further validating its effectiveness in modeling both object detection and association.

Taken together, the results across both datasets clearly demonstrate that GL-DT not only enhances detection quality but also effectively mitigates performance bottlenecks in complex scenarios. It consistently delivers systemic advantages in identity preservation, tracking accuracy, and robustness. More importantly, GL-DT achieves these gains while maintaining real-time efficiency, underscoring its broad applicability and high value as a state-of-the-art multi-object tracking method.

\subsection{Ablation Experiment}
\subsubsection{Detection Component}

\begin{figure}[htbp]
    \centering
    \includegraphics[width=1\linewidth]{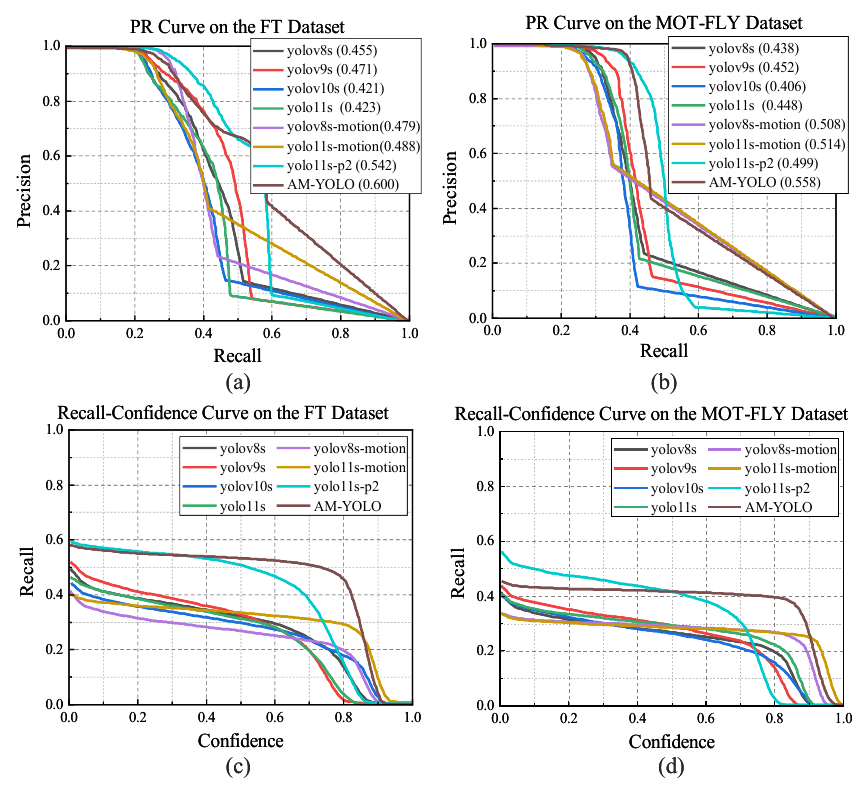}
    \caption{PR Curve and Recall-Confidence Curve on the FT and MOT-FLY Dataset.}
    \label{Fig7}
\end{figure}

Within the tracking-by-detection paradigm, the accuracy of object detection is the key factor determining the reliability of multi-object tracking for UAVs. The proposed AM-YOLO model significantly enhances the detection performance of small UAV targets. To validate its effectiveness, we conducted comparative experiments against several mainstream YOLO models (YOLOv8s, YOLOv9s, YOLOv10s, YOLOv11s), and evaluated their detection performance across different datasets using Precision–Recall (PR) curves, the corresponding area under the curve (AUC), and Recall–Confidence curves, as illustrated in Fig. \ref{Fig7}.

\begin{table*}[htbp]
  \centering
  \caption{Impact of JCMA and PMR on Tracking Performance Across Two Datasets.}
  \begin{tabular}{c|c|c|c|c|c|c|c|c}
    \hline
    \textbf{Dataset} & Model &JCMA &PMR & IDSW$\downarrow$ & IDF1$\uparrow$ & MOTA$\uparrow$ & MOTP$\uparrow$ & HOTA$\uparrow$ \\
    \hline
    \multirow{3}{*}{\centering MOT-FLY Dataset} 
    & AM-YOLO + ByteTrack & & & 92 & 62.01 & 73.89 & 82.62 & 53.16 \\
    & AM-YOLO + JCMA & \checkmark & & 44 & 74.82 & 76.45 & 82.41 & 61.62 \\
    & AM-YOLO + JCMA + PMR & \checkmark & \checkmark & 43 & 76.40 & 76.41 & 82.42 & 62.12 \\
    \hline
    \multirow{3}{*}{\centering FT Dataset} 
    & AM-YOLO + ByteTrack & & & 84 & 32.83 & 50.69 & 73.26 & 35.28 \\
    & AM-YOLO + JCMA & \checkmark & & 27 & 56.43 & 54.22 & 72.56 & 46.02 \\
    & AM-YOLO + JCMA + PMR & \checkmark  & \checkmark  & 23 & 57.33 & 54.27 & 72.55 & 46.69 \\
    \hline
  \end{tabular}
  \label{tab3}
\end{table*}

\begin{table*}[h]
\centering
\caption{Effect of combining global and local detection on multi-object tracking performance}
\begin{tabular}{c|c|c|c|c|c|c|c|c}
\hline
\textbf{Dataset} & \textbf{Model} & GD  & LD  & \textbf{IDSW↓} & \textbf{IDF1↑} & \textbf{MOTA↑} & \textbf{MOTP↑} & \textbf{HOTA↑} \\
\hline
\multirow{4}{*}{\centering MOT-FLY Dataset} 
& GD + ByteTrack &\checkmark & & 92 & 62.01 & 73.89 & 82.62 & 53.16 \\
& GD + LD + ByteTrack & \checkmark &  \checkmark & 72 & 69.11 & 78.27 & 82.64 & 58.73 \\
& GD + JPTrack &\checkmark & & 43 & 76.40 & 76.41 & 82.42 & 62.12 \\
& GD + LD + JPTrack (Ours)  & \checkmark &  \checkmark & 33 & 79.21 & 82.84 & 82.11 & 65.22 \\
\hline
\multirow{4}{*}{FT-Dataset} 
& GD + ByteTrack &\checkmark & & 84 & 32.83 & 50.69 & 73.26 & 35.28 \\
& GD + LD + ByteTrack  & \checkmark &  \checkmark & 51 & 37.73 & 61.78 & 76.07 & 37.28 \\
& GD + JPTrack &\checkmark & & 23 & 57.33 & 54.27 & 72.55 & 46.69 \\
& GD + LD + JPTrack (Ours)  & \checkmark &  \checkmark & 14 & 83.99 & 72.06 & 74.76 & 60.73 \\
\hline
\end{tabular}
\label{tab4}
\end{table*}

From the PR curves shown in Fig. \ref{Fig7} (a) and (b), it can be observed that UAV targets are mostly small-scale objects, and traditional three-detection-head structures (YOLOv8s–YOLOv11s) exhibit limited performance on such targets. The YOLO11s-p2 model, specifically designed for small objects, achieved AUC values of 0.499 and 0.542 on the MOT-FLY and FT datasets, respectively, outperforming the baseline models. Additionally, the YOLOv8s-motion and YOLO11s-motion models, which integrate appearance and motion information, demonstrated improved accuracy compared to models relying solely on appearance features, indicating the positive role of motion features in detection. The proposed AM-YOLO model, which combines a small-object detection head with an appearance–motion feature fusion mechanism, achieved the best performance on the PR curves of both datasets, with AUC values increasing to 0.558 and 0.600, respectively. This represents an improvement of 0.197 and 0.295 over YOLOv11s, validating its effectiveness and superiority in UAV small-object detection tasks.

Furthermore, from the Recall–Confidence curves shown in Fig. \ref{Fig7} (c) and (d), it can be seen that AM-YOLO maintains the highest recall rate across most confidence intervals. This indicates that, at the same confidence threshold, the model can detect more true targets, reflecting its excellent generalization ability and confidence calibration quality. This characteristic demonstrates that the confidence scores of the detection boxes output by AM-YOLO are more reliable, which helps in setting appropriate thresholds in practical applications. By suppressing false detections while maintaining a high target recall rate, the overall robustness of the tracking system is enhanced.

\subsubsection{Tracking Component}

To validate the effectiveness of the JCMA and PRM modules in the proposed tracking method, this study analyzes the impact of introducing different modules on multi-object tracking performance while keeping the detector unchanged. The results are presented in Table~\ref{tab3}. Specifically, AM-YOLO + ByteTrack was designated as the baseline model, and evaluations were carried out on the MOT-FLY and FT datasets.

Compared with the baseline, incorporating JCMA yielded substantial improvements across both datasets. On the MOT-FLY dataset, IDSW dropped from 92 to 44, IDF1 increased from 62.01 to 74.82, MOTA improved by 3.34\%, and HOTA rose by 13.7\%. On the FT dataset, IDSW decreased sharply from 84 to 27, while IDF1 improved from 32.83 to 56.43, MOTA increased by 6.5\%, and HOTA achieved a remarkable gain of 23.33\%. These results highlight JCMA’s pivotal role in reducing identity switches and enhancing association accuracy, thereby strengthening consistency modeling under identical detection conditions.

Building on this, the introduction of the PMR module further optimized performance. On the MOT-FLY dataset, IDSW dropped to 43, IDF1 increased to 76.40, and HOTA rose to 62.12. On the FT dataset, IDSW was further reduced to 23, with IDF1 increasing to 57.33 and HOTA reaching 46.69. These outcomes demonstrate that PMR enhances tracking continuity and consistency while further mitigating the risk of identity switches. In summary, the integration of JCMA and PMR effectively boosts tracking performance, and their synergistic contribution enables optimal multi-object tracking outcomes.

\subsubsection{Global-Local Detection Component}
To evaluate the impact of combining global and local detectors on multi-object tracking performance, we conducted systematic experiments, with results summarized in Table~\ref{tab4}. Here, GD denotes the Global Detection and LD the Local Detection.

The results indicate that using the global detector alone with ByteTrack (GD + ByteTrack) yields the poorest tracking performance, characterized by frequent ID switches (highest IDSW) and relatively low IDF1, MOTA, and HOTA scores. Introducing the local detector to form a GD + LD joint detection framework significantly improves performance: IDSW decreases notably, while IDF1, MOTA, and HOTA all increase substantially. Further integrating the JPTrack module, the GD + LD + JPTrack architecture demonstrates additional gains over GD + JPTrack alone. On the MOT-FLY dataset, IDSW decreases from 43 to 33, IDF1 improves by 2.81\%, MOTA by 6.43\%, aFT dataset, IDSW drops from 23 to 14, IDF1 rises by 26.66\%, MOTA by 17.79\%, and HOTA by 14.04\%.

\begin{figure*}[htbp]
    \centering
    \setlength{\tabcolsep}{2pt}  
    \renewcommand{\arraystretch}{1.5}  
    \begin{tabular}{cccc}  
        \includegraphics[width=0.24\textwidth]{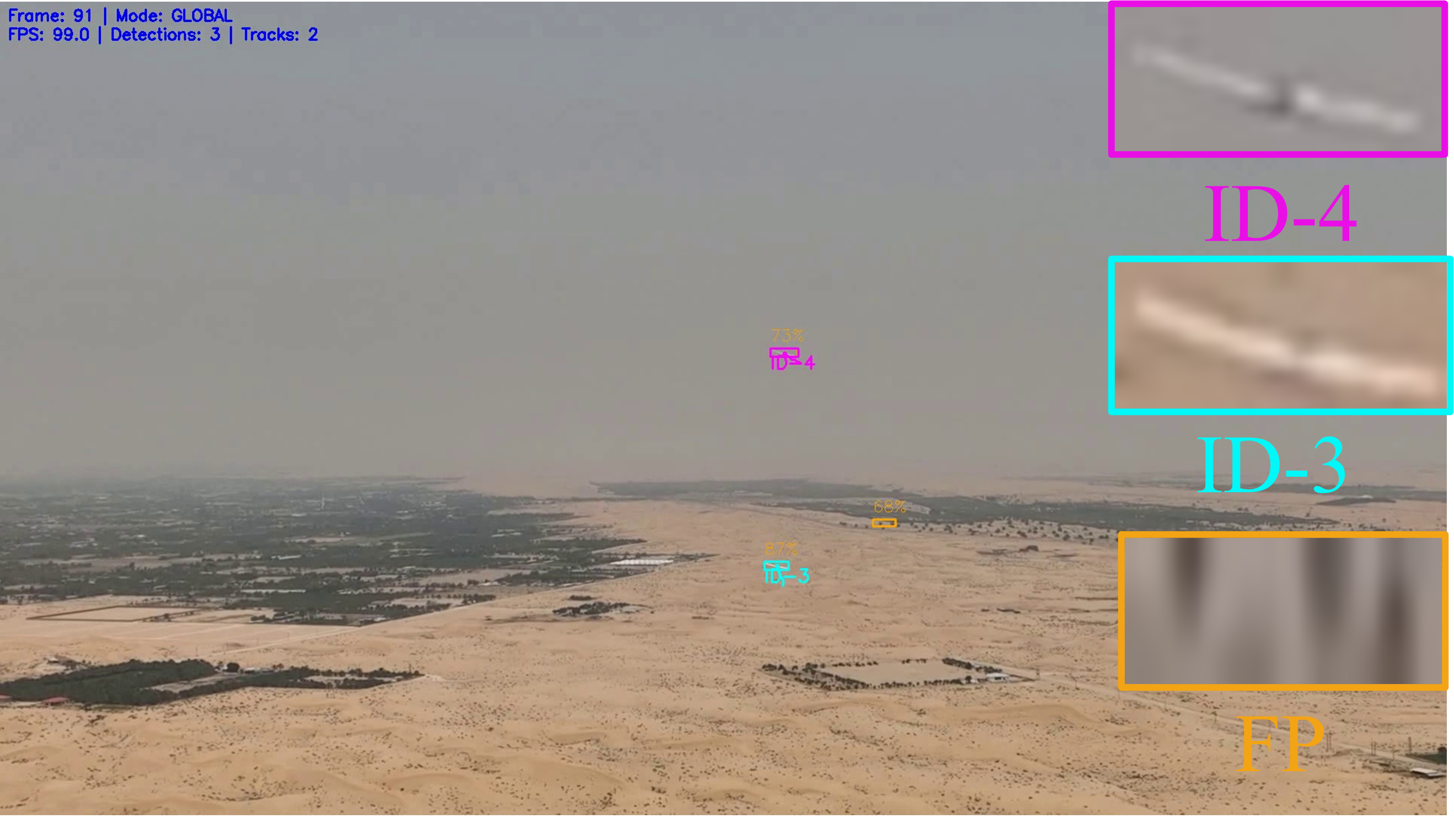} &
        \includegraphics[width=0.24\textwidth]{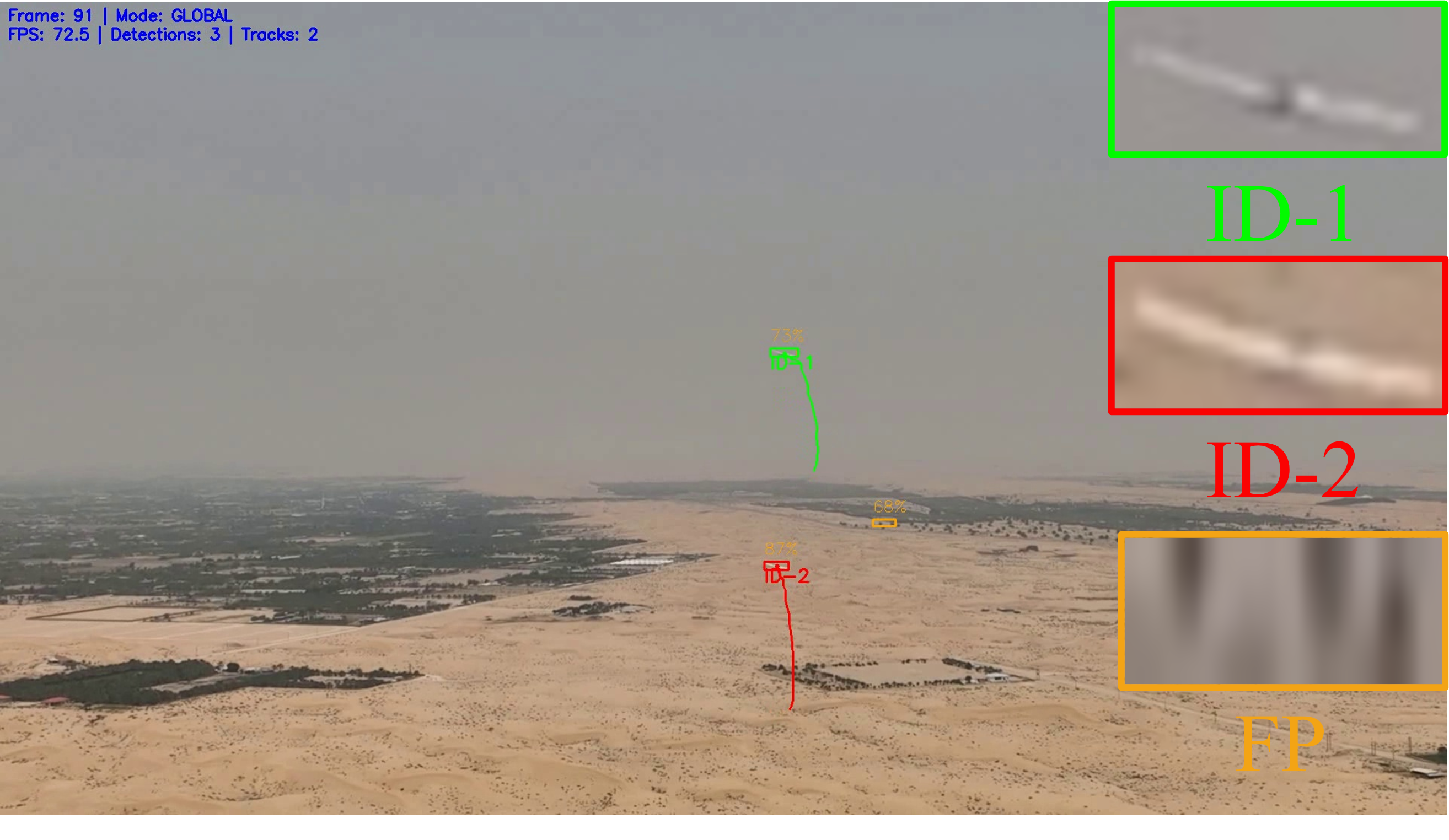} &
        \includegraphics[width=0.24\textwidth]{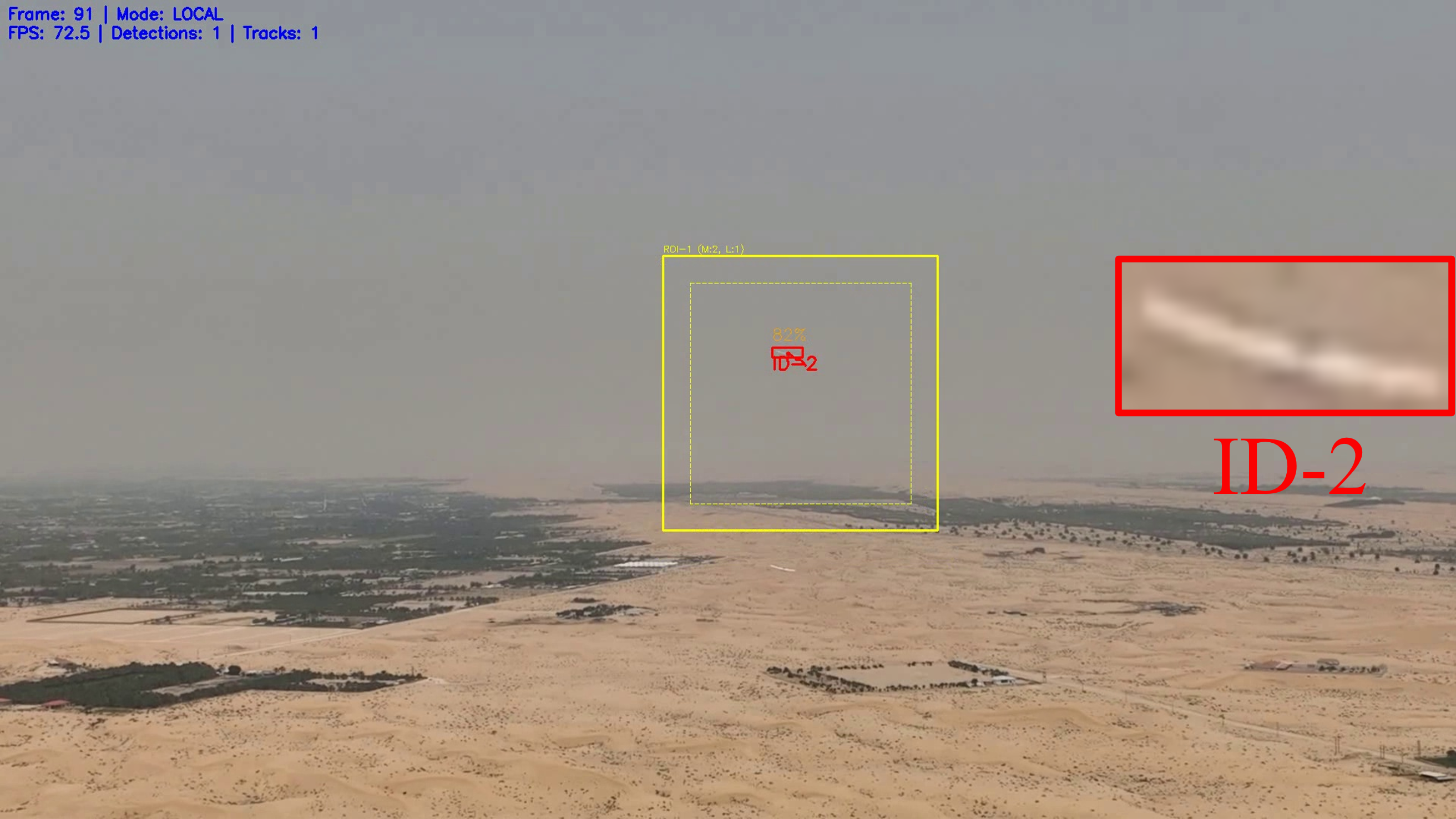} &
        \includegraphics[width=0.24\textwidth]{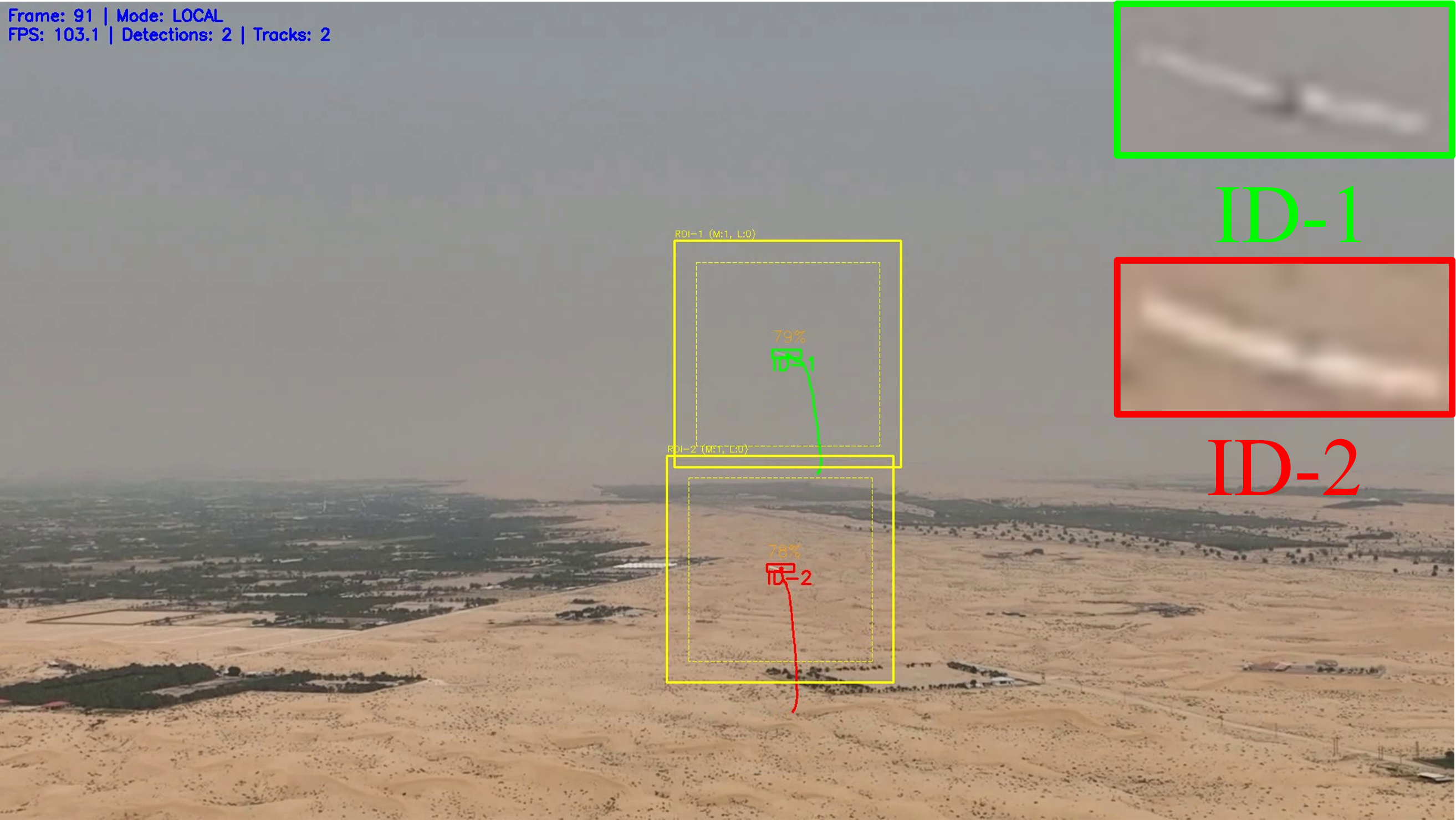} \\
        
        \includegraphics[width=0.24\textwidth]{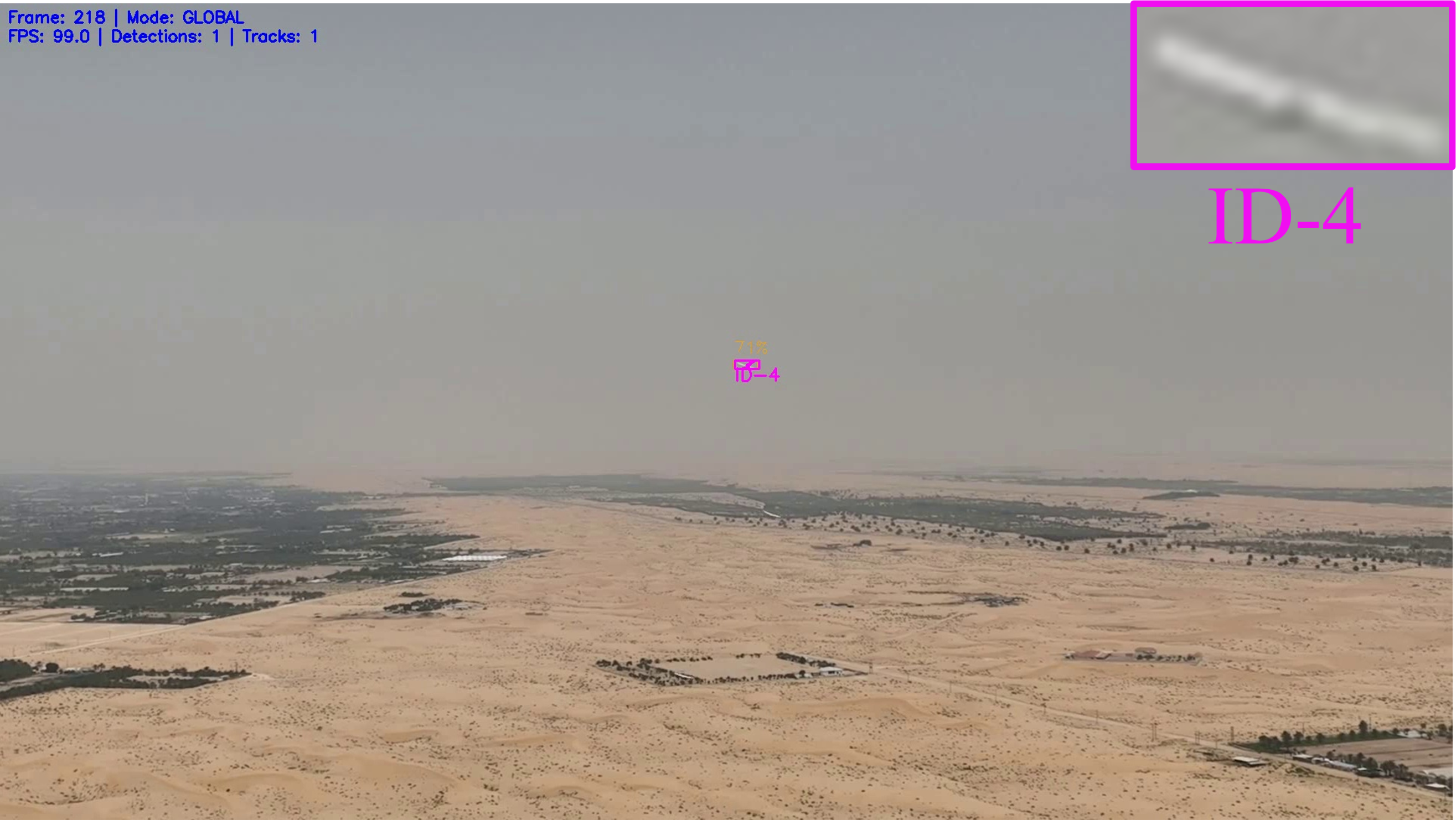} &
        \includegraphics[width=0.24\textwidth]{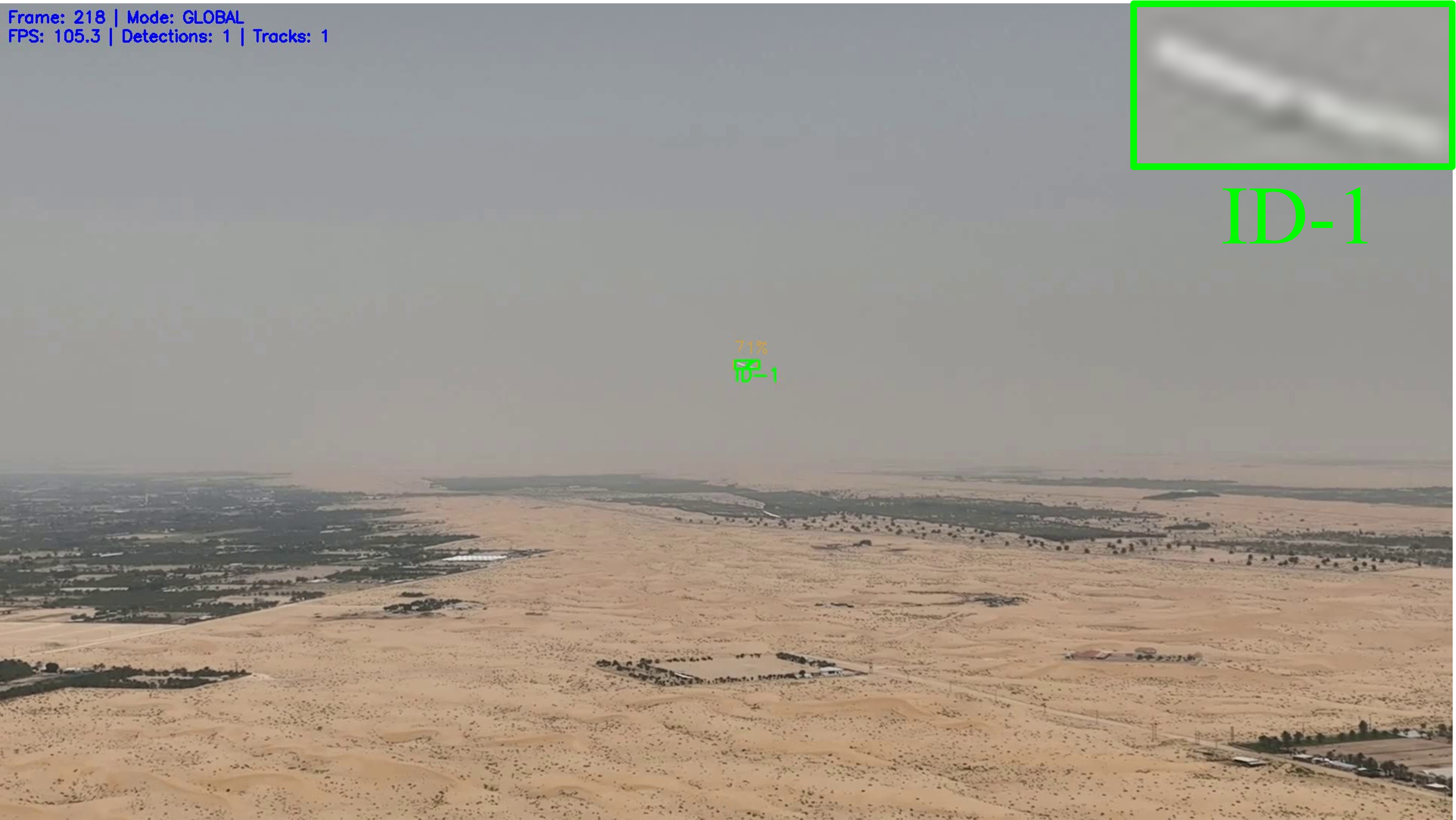} &
        \includegraphics[width=0.24\textwidth]{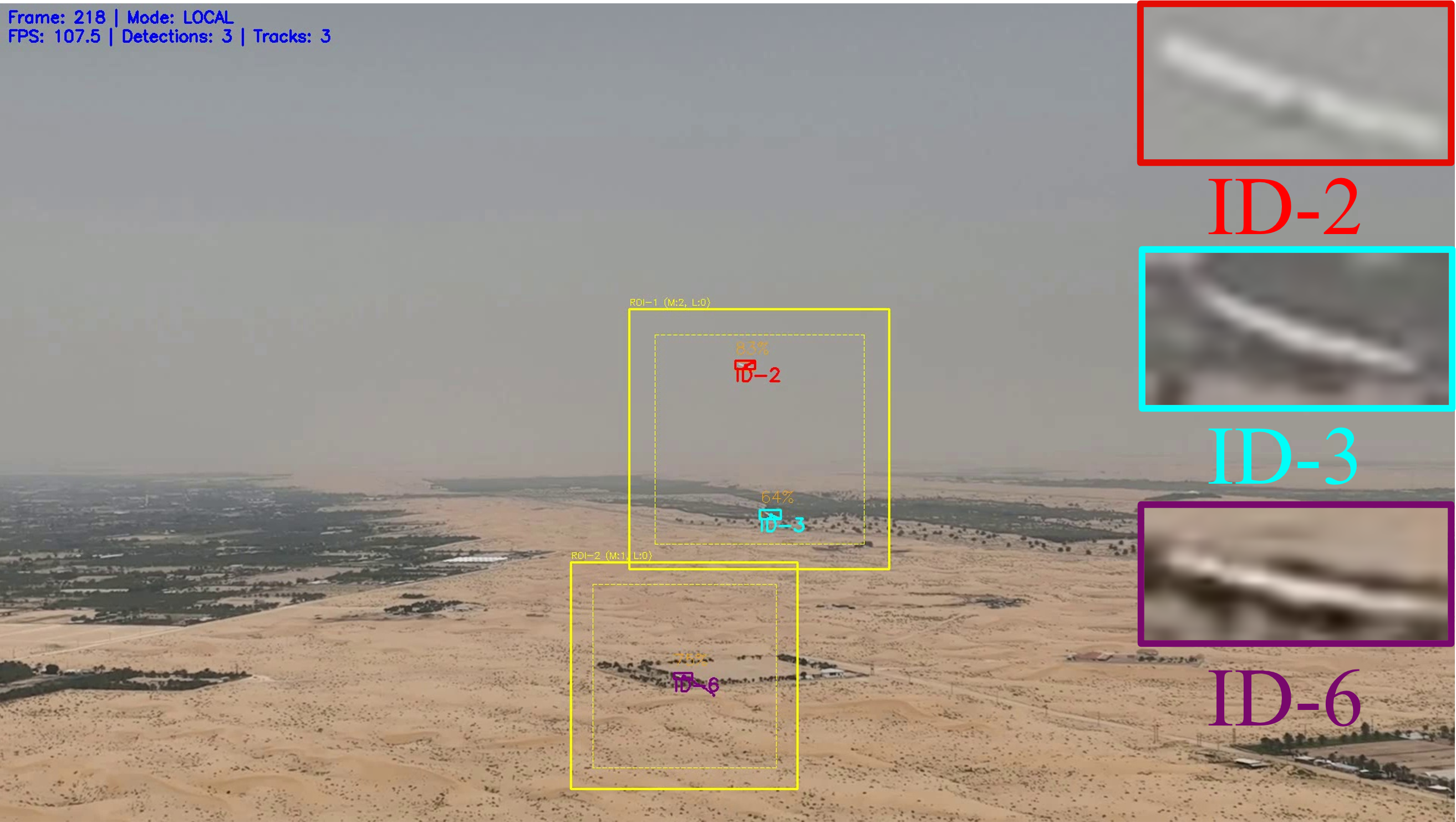} &
        \includegraphics[width=0.24\textwidth]{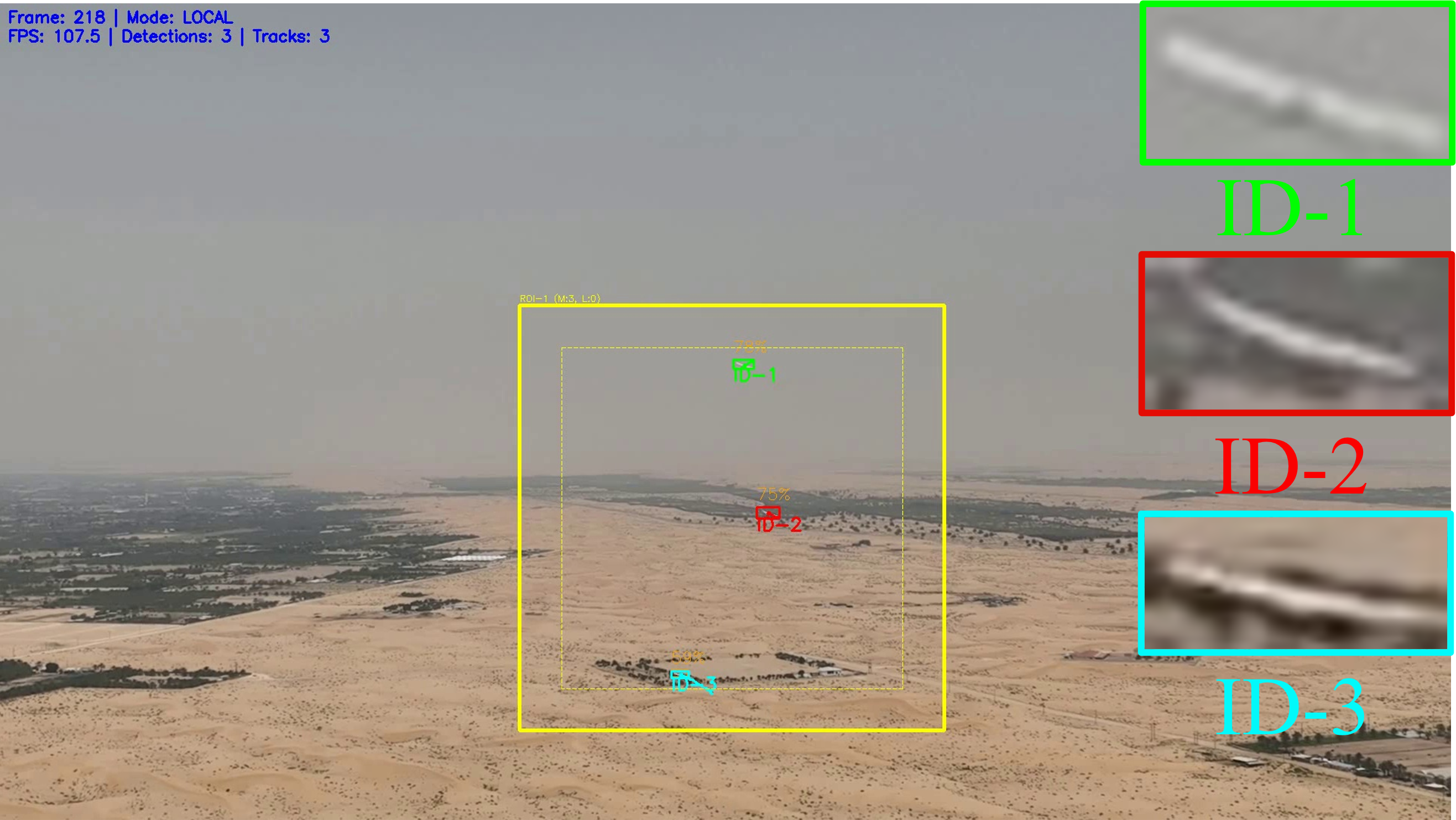} \\
        
        \includegraphics[width=0.24\textwidth]{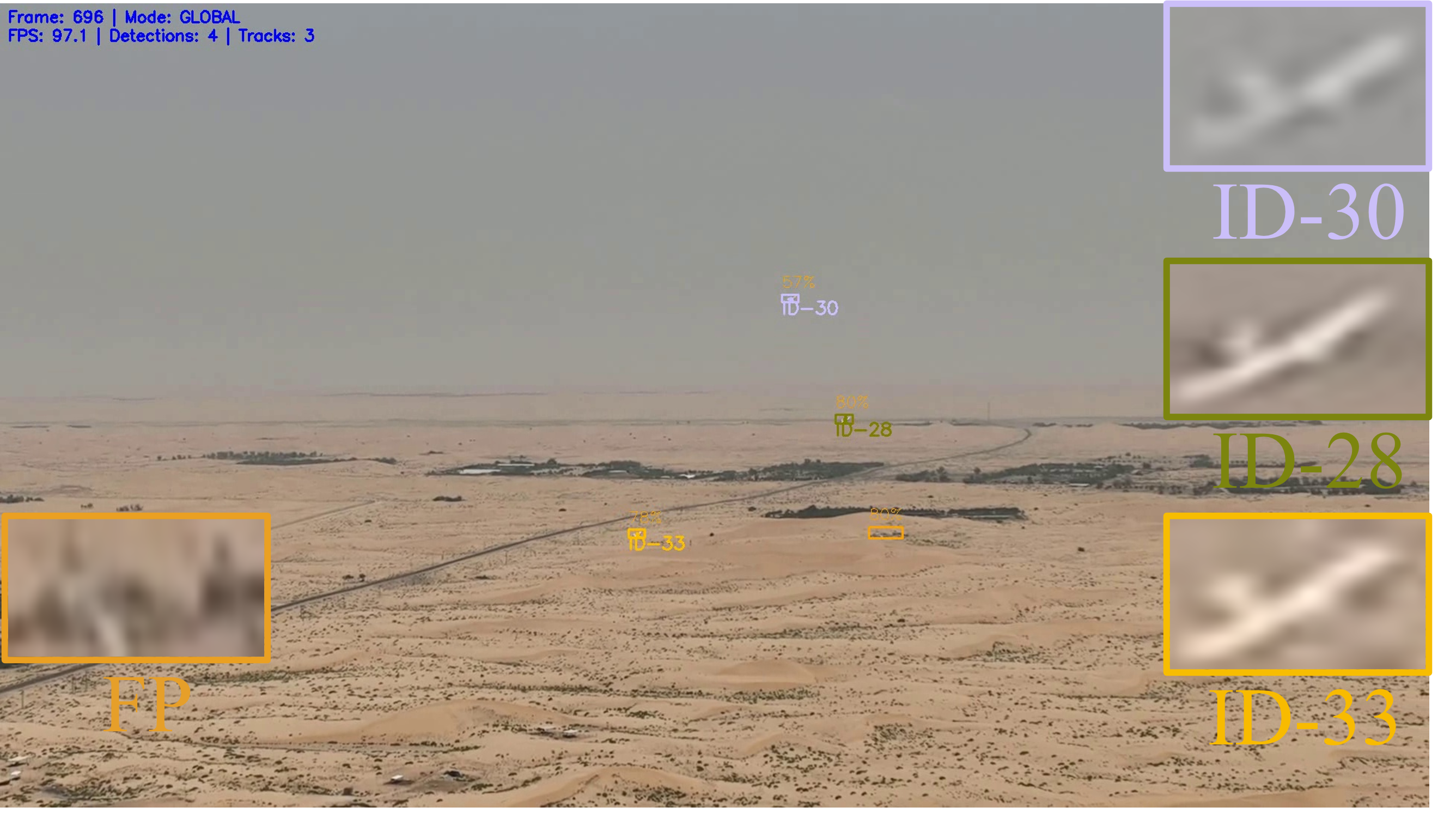} &
        \includegraphics[width=0.24\textwidth]{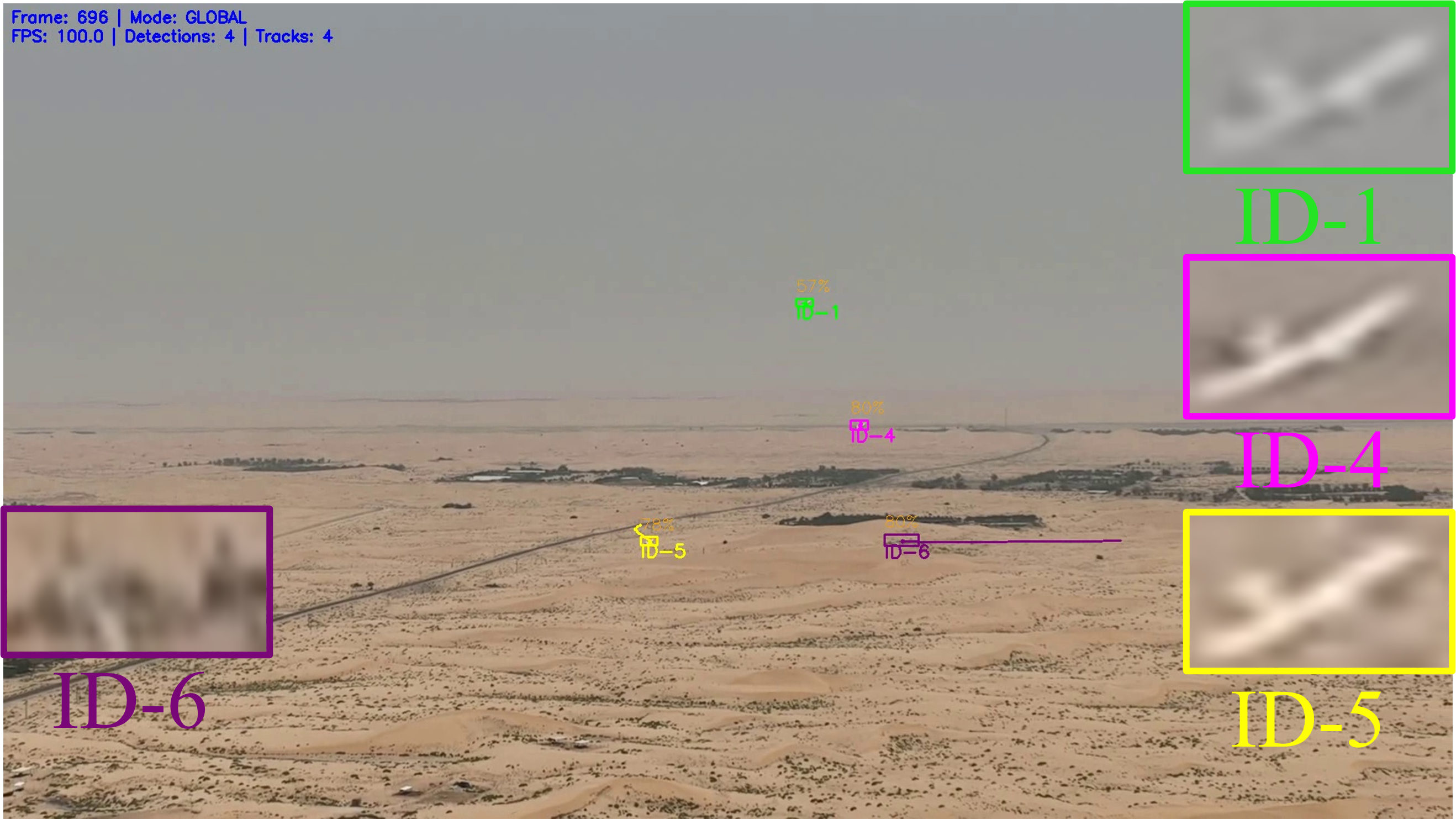} &
        \includegraphics[width=0.24\textwidth]{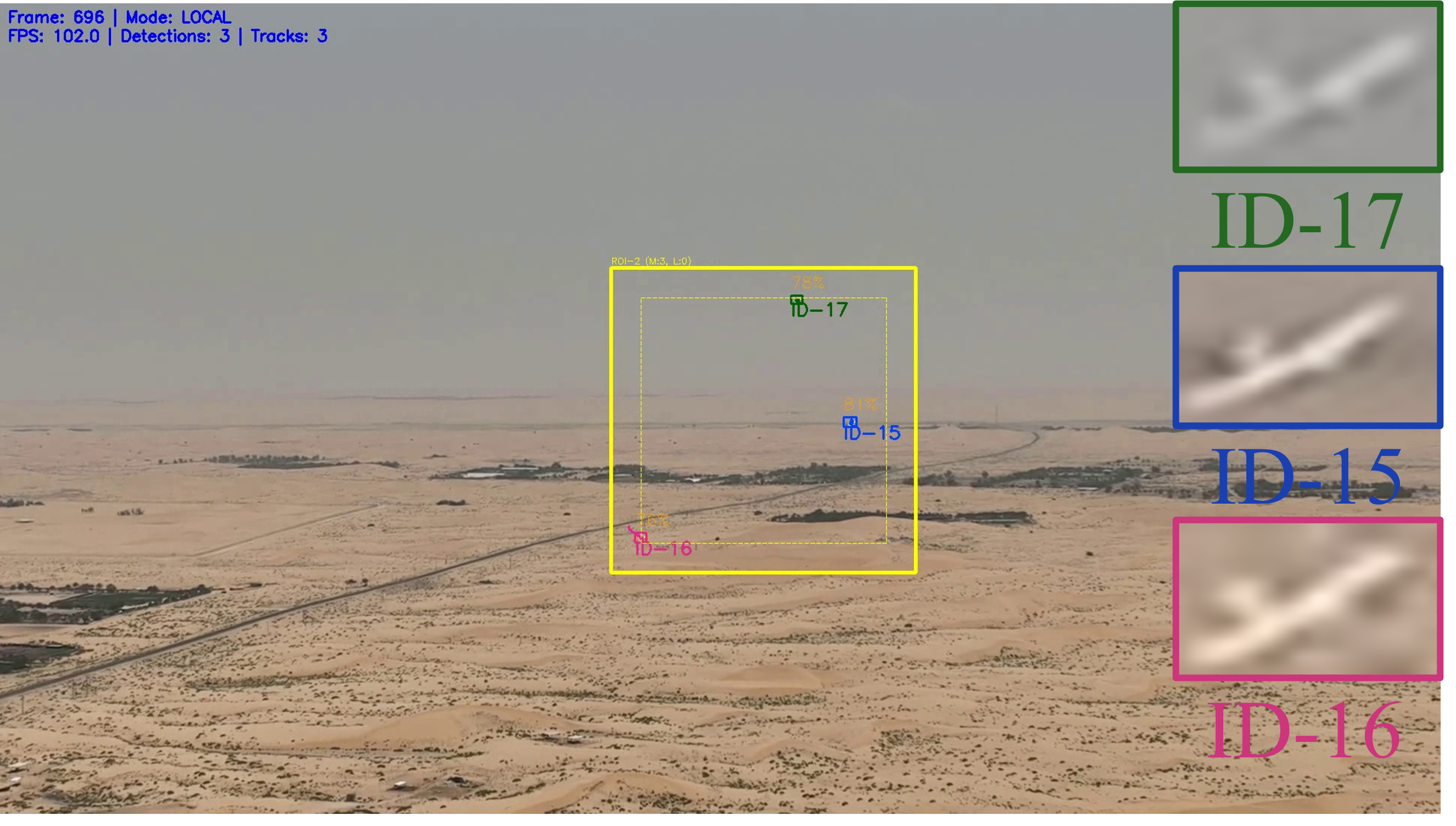} &
        \includegraphics[width=0.24\textwidth]{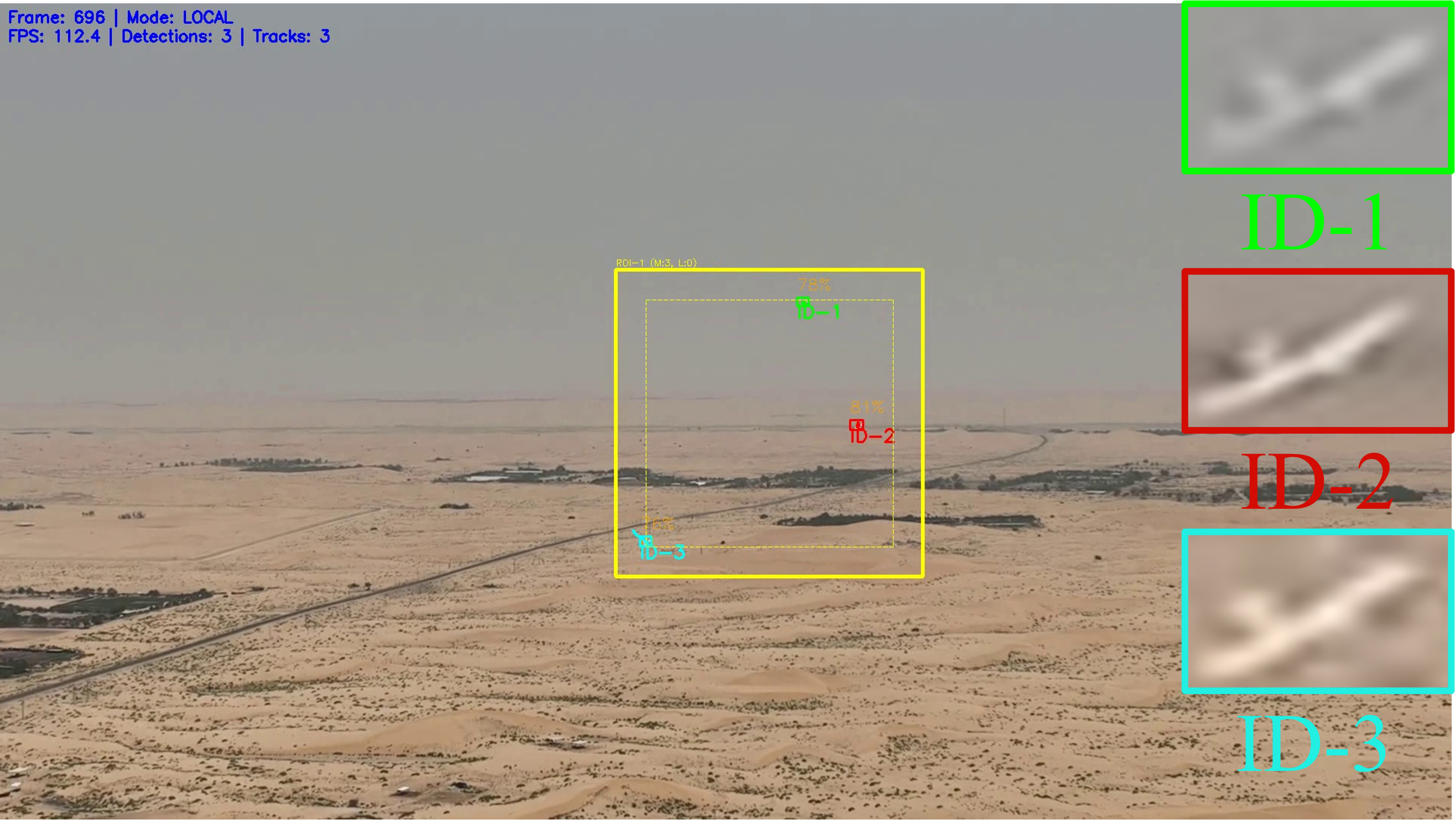} \\
        
        \includegraphics[width=0.24\textwidth]{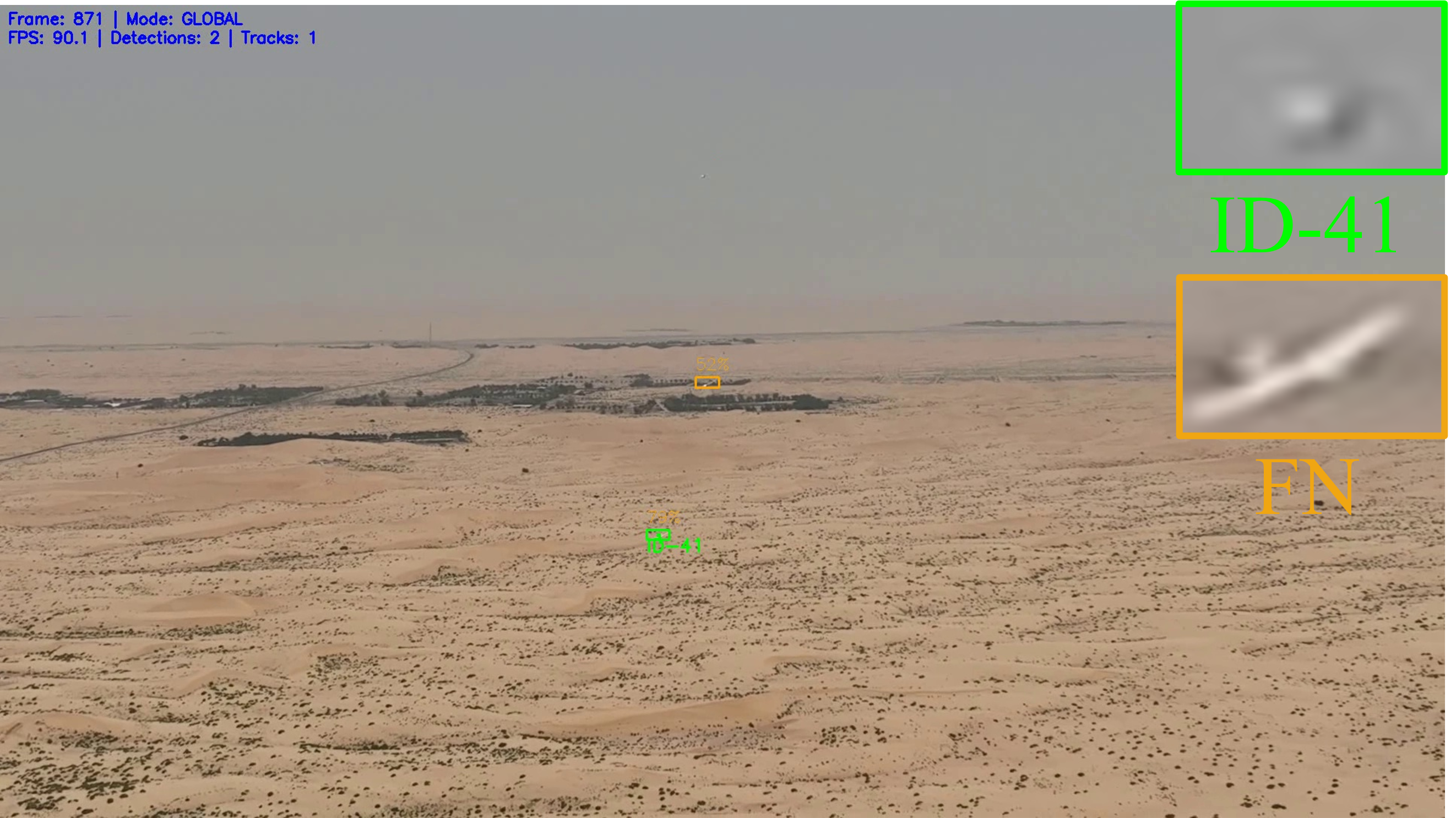} &
        \includegraphics[width=0.24\textwidth]{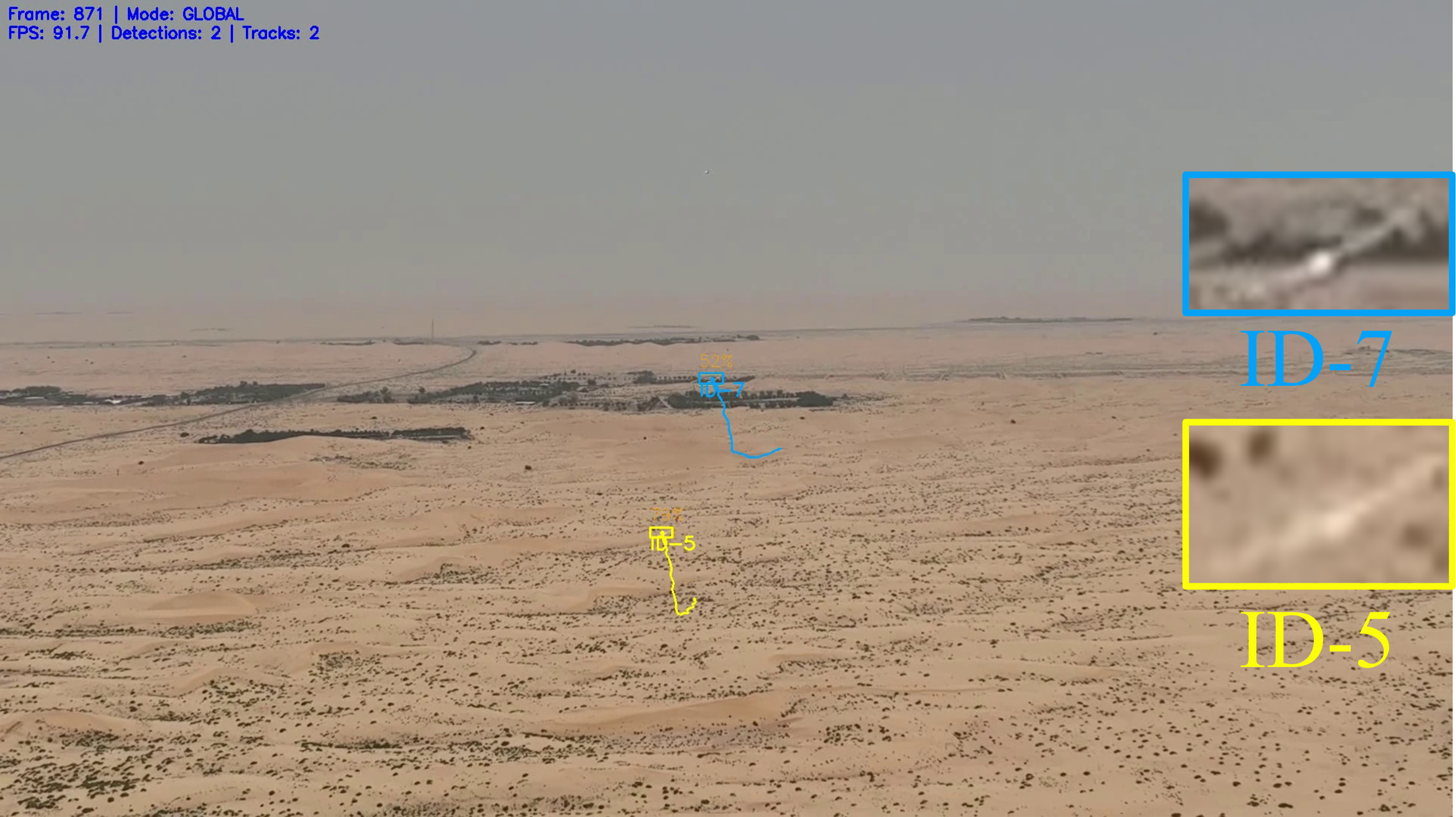} &
        \includegraphics[width=0.24\textwidth]{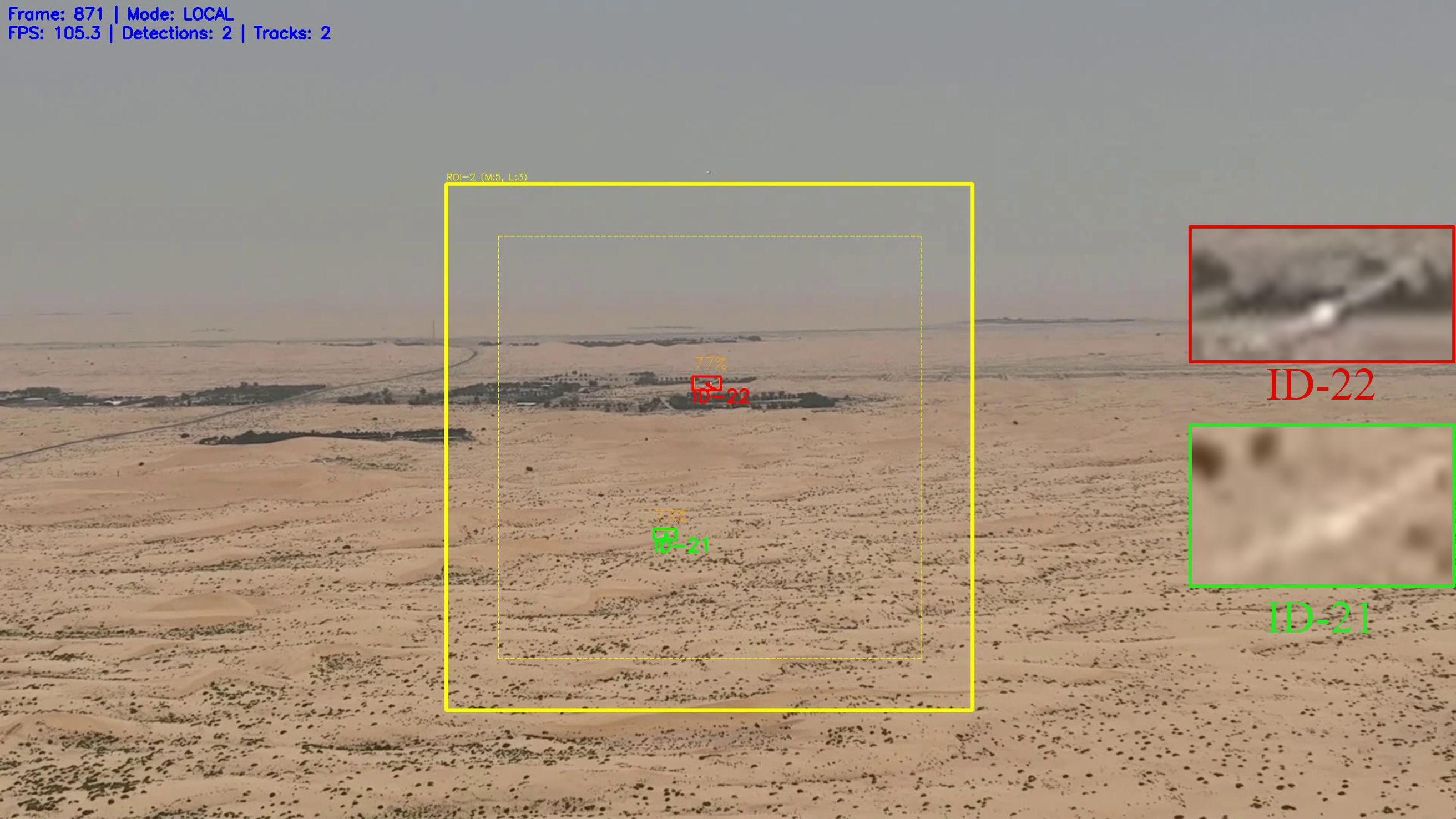} &
        \includegraphics[width=0.24\textwidth]{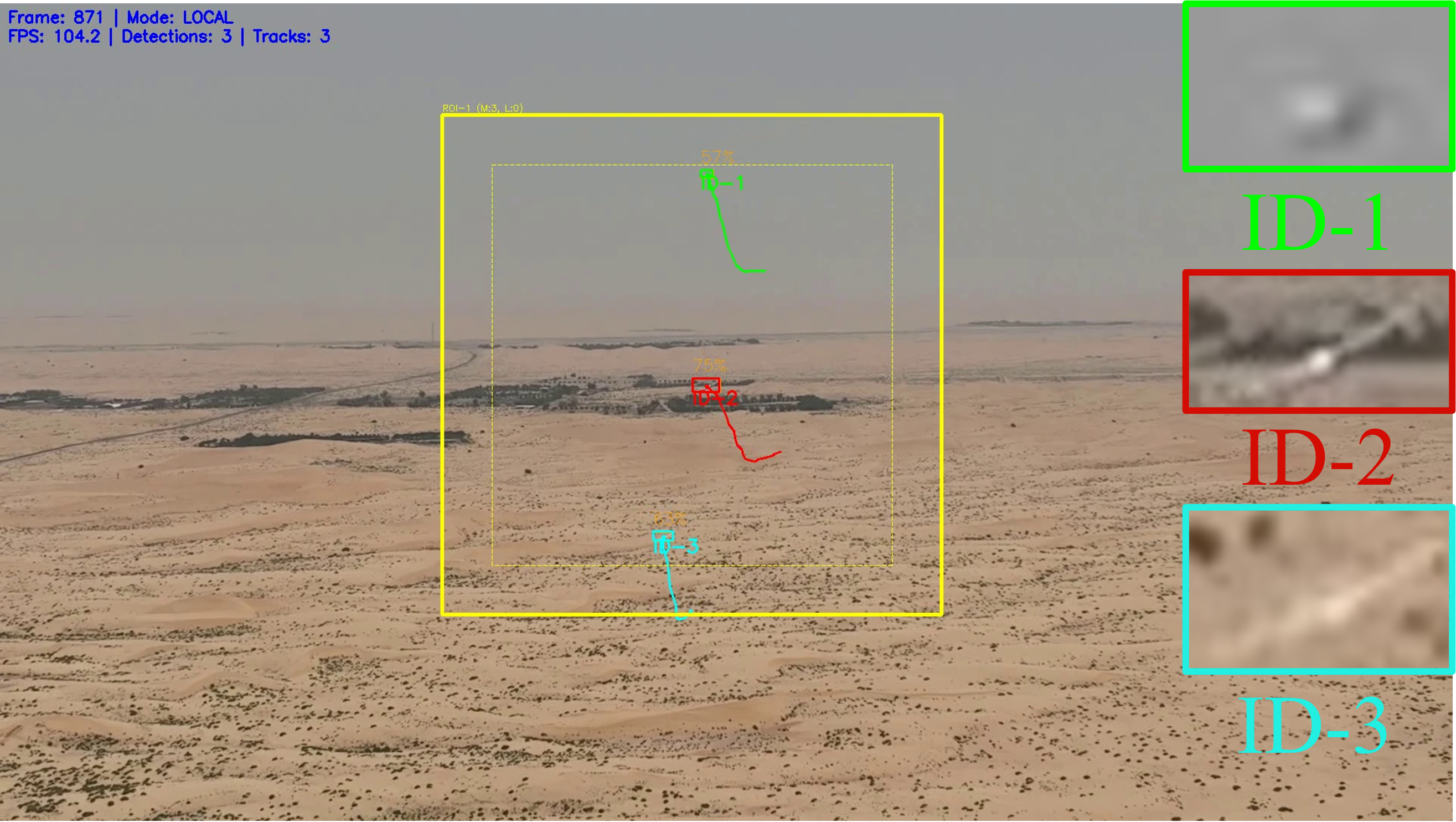} \\
        
        \multicolumn{1}{c}{\small{(a) GD + ByteTrack}} &
        \multicolumn{1}{c}{\small{(b) GD + JPTrack}} &
        \multicolumn{1}{c}{\small{(c) GD + LD + ByteTrack}} &
        \multicolumn{1}{c}{\small{(d) GD + LD + JPTrack (Ours) }} \\
    \end{tabular}
    
    \caption{Representative tracking results on the FT dataset using four methods: (a) GD+ByteTrack, (b) GD+JPTrack, (c) GD+LD+ByteTrack, and (d) GD+LD+JPTrack (ours). FP and FN denote false positives and false negatives, respectively. Our GD+LD+JPTrack approach achieves the best overall performance, delivering the most stable tracking results across the three UAVs.}
    \label{Fig8}
\end{figure*}

This performance improvement is primarily attributed to the small size of UAV targets, where a single global detector struggles to detect small objects in complex backgrounds. The GD + LD + JPTrack framework leverages the global detector’s full-scene awareness and the local detector’s fine-grained small-object recognition, combined with an optimized tracking module, to enhance trajectory continuity and stability. All metrics reach optimal levels, demonstrating the synergistic effect between joint detection and optimized tracking, significantly improving both accuracy and robustness. The results validate the effectiveness and necessity of the proposed global-local detection framework for UAV multi-object tracking tasks. Fig. \ref{Fig8} presents several representative examples, showcasing the multi-object tracking performance achieved using different methods. It can be seen from the figure that the GD+LD+JPTrack method we proposed has achieved the best performance.

\subsubsection{Inference Time}

Considering the computational constraints in real-world applications, the proposed algorithm was further optimized for efficient inference and embedded deployment. Specifically, the model was accelerated using the TensorRT inference engine, with CUDA-based parallelization applied to both preprocessing and post-processing stages to enhance overall efficiency. In a PC environment, the original Python implementation achieved approximately 35.88 FPS, while the optimized version reached an average of 124 FPS. Furthermore, when deployed on the embedded computing platform NVIDIA Jetson Xavier NX, the model achieved a practical inference speed of 25 FPS, thereby satisfying the requirement for real-time performance.

\section{CONCLUSION}
This paper presents a real-time multi-object detection and tracking framework for UAVs, designed to address the challenges of small target sizes, frequent occlusions, and complex multi-target interactions in dynamic environments. By incorporating the STFF module and a global–local collaborative detection strategy, the proposed system significantly improves small-object recognition while effectively suppressing background interference. In addition, the integration of the JPTrack tracking algorithm ensures continuous and stable trajectory estimation for multiple targets. Experimental results confirm that the STFF module enhances the perception of both motion and appearance features, thereby improving detection accuracy; the global–local collaboration strategy further strengthens the recognition of small targets. Moreover, JPTrack’s three-stage matching mechanism achieves higher matching success rates, demonstrating superior performance in identity preservation and tracking accuracy. Overall, GL-DT outperforms state-of-the-art methods on the MOT-FLY and FT datasets, highlighting its effectiveness and practical value for UAV-based multi-object detection and tracking.

In summary, GL-DT provides a feasible and efficient framework for multi-object tracking with UAVs, demonstrating strong practical potential in real-world aerial surveillance and monitoring tasks. Future research will focus on developing more efficient global–local coordination mechanisms and enhancing feature extraction for high-speed and emergent targets, thereby enabling more accurate pose perception and sustained, stable tracking.

\bibliographystyle{IEEEtran}
\bibliography{references}

\end{document}